\documentclass{article}

\usepackage{microtype}
\usepackage{graphicx}
\usepackage{booktabs}
\usepackage{xurl}
\usepackage{hyperref}
\usepackage{algorithm}
\usepackage{algorithmic}

\newcommand{\algheader}[2][]{%
  \par\medskip\refstepcounter{algorithm}%
  \ifx\\#1\\\else\label{#1}\fi%
  \noindent\textbf{Algorithm~\thealgorithm.}~\emph{#2}\par\smallskip%
}

\usepackage[preprint]{neurips_2026}

\usepackage{amsmath, amssymb}
\usepackage{tikz}
\usetikzlibrary{shapes.geometric, arrows.meta, positioning, fit, backgrounds, calc}
\usepackage{pgfplots}
\pgfplotsset{compat=1.18}
\usepackage{pifont}
\newcommand{\cmark}{\ding{51}}
\newcommand{\xmark}{\ding{55}}
\usepackage{multirow}
\usepackage{enumitem}
\usepackage{listings}
\usepackage{caption}
\usepackage{placeins}
\usepackage{xcolor}

\newcommand{\figcaption}[1]{\vspace*{1mm}\caption{#1}\vspace*{-2mm}}

\newcommand{\ignore}[1]{}

\makeatletter
\renewcommand{\section}{\@startsection{section}{1}{\z@}%
  {-1.6ex plus -0.4ex minus -0.2ex}%
  {0.8ex plus 0.2ex}%
  {\normalfont\large\bfseries}}
\renewcommand{\subsection}{\@startsection{subsection}{2}{\z@}%
  {-1.2ex plus -0.3ex minus -0.2ex}%
  {0.5ex plus 0.1ex}%
  {\normalfont\normalsize\bfseries}}
\renewcommand{\paragraph}{\@startsection{paragraph}{4}{\z@}%
  {0.7ex plus 0.2ex minus 0.1ex}%
  {-0.7em}%
  {\normalfont\normalsize\bfseries}}
\makeatother
\setlist{nosep,leftmargin=*}

\begin{document}

\title{Atomix: Timely, Transactional Tool Use for Reliable Agentic Workflows}

\author{%
  Bardia Mohammadi\textsuperscript{1},
  Nearchos Potamitis\textsuperscript{2},
  Lars Klein\textsuperscript{3},\\[0.7ex]
  \textbf{Akhil Arora\textsuperscript{2}},
  \textbf{Laurent Bindschaedler\textsuperscript{1}}\\[1.5ex]
  \textsuperscript{1}Max Planck Institute for Software Systems\quad
  \textsuperscript{2}Aarhus University\quad
  \textsuperscript{3}EPFL\\[0.9ex]
  \texttt{\{bmohammadi, bindsch\}@mpi-sws.org}
}

\maketitle

\begin{abstract}
LLM agents execute multi-step workflows that mutate external state through tools. Common orchestrators treat tool return as the settlement trigger, so faults, speculation, and concurrent agents can leave partial effects, losing-branch residue, stale writes, or irreversible sends. Correct settlement needs two facts that retries, checkpoint replay, locks, and compensation each conflate: which effects must settle together, and when earlier conflicting work is exhausted. \textbf{Atomix} makes this split explicit with progress-aware transactions. The runtime records reads and effects during execution, seals a transaction when its footprint is complete, and commits only after per-resource frontiers show that no earlier conflicting work can still arrive. Commit is final settlement: Atomix releases bufferable effects, accepts reversible external effects as final, and lets irreversible effects leave the gate. Abort suppresses unreleased effects and compensates externalized reversible effects where possible. On representative agent workloads, this composition improves clean recovery under injected faults, isolates contending and speculative work, and prevents correctly classified irreversible actions from leaking; microbenchmarks show microsecond-scale wrapper overhead relative to tool latency. The code and artifacts of Atomix are available at \url{https://github.com/mpi-dsg/atomix}.
\end{abstract}

% Section break

\section{Introduction}
\label{sec:introduction}

LLM agents execute multi-step tasks that mutate external state: editing code, updating databases, filling web forms, sending emails, and calling third-party APIs~\cite{langgraph_docs,crewai_docs,wu2023autogen,claude_code_docs,openai_function_calling_guide,mcp_spec_2025}. These workloads increasingly use concurrent execution: multiple agents share resources, parallel plans explore alternatives, and speculative execution starts downstream work before upstream confirmation. Each pattern raises the same question: \emph{when may a tool effect become permanent?}

Current agent frameworks expose this decision indirectly. Orchestrators provide checkpointing, retries, and timeouts, and rely on tool-level idempotency keys plus Saga-style compensations~\cite{langgraph_docs,temporal_docs,aws_step_functions,garcia1987sagas}. These mechanisms recover after a tool effect has externalized or after the workflow restores its own state. That execution model leaves three gaps once workflows run in parallel:
\begin{enumerate}[leftmargin=*, nosep]
\item \emph{Speculation}: parallel plans execute real tool calls; losing branches leave residual side effects.
\item \emph{Contention}: concurrent agents touch the same resource and interleave writes.
\item \emph{Irreversibility}: some effects cannot be undone once externalized.
\end{enumerate}

\textbf{Atomix} adds transactional settlement at the tool interface. Atomix separates execution, sealing, frontier checks, and settlement. During execution, adapters record the resources a transaction reads and the effects it may produce. The orchestrator then \emph{seals} the transaction, declaring that no more reads, scopes, or effects will be added. Atomix commits a sealed transaction only when per-resource frontiers certify, under the orchestrator's advancement rule, that no earlier work on any touched resource can still arrive. Commit settles the transaction: Atomix releases staged effects, lets irreversible effects leave the gate, and accepts already-externalized reversible effects as final. Abort suppresses unreleased effects and compensates externalized reversible effects where possible. The design adapts Sagas, Try-Confirm-Cancel, and streaming watermarks to the LLM tool interface. It requires orchestrator cooperation for progress tracking, but tools need no modification when all calls pass through adapters.

\paragraph{Why This Design Is Needed.}
No single existing trigger both admits speculative execution and decides when effects may safely become external. If tool return triggers settlement, a losing speculative branch or a failed later step has already mutated the world. If checkpoint rollback is the recovery mechanism, the runtime restores agent memory but not an email, booking, or remote API update. Saga compensation helps only when every externalized effect is reversible and has a correct handler; it cannot prevent an irreversible send. Locks cover only part of the problem: per-call locks miss the read--plan--write gap, while workflow-wide locks serialize unrelated work. Optimistic revalidation detects some stale writes, but still needs a closed transaction footprint and a rule for when irreversible effects may leave the gate. Frontiers provide ordering, but they do not make effects atomic. Transactions group effects, but by themselves do not show that earlier conflicting work has finished. Atomix therefore combines transaction grouping, seal, per-resource progress, and effect-class settlement.

\paragraph{Evaluation.}
We evaluate fault recovery, frontier-gated isolation, and irreversible-effect gating, then combine them in a stress test. Under heavy fault injection on $\tau$-bench, Atomix retains 57\% clean task success while every non-Checkpoint-Replay baseline falls to 0--7\%. Under forced multi-agent overlap it records 0 conflict-cycle witnesses. On irreversible sends it leaks 0/500 invalid sends and releases all 500 valid ones. On combined stress, Atomix is Pareto-best on the joint surface: in the top run-clean tier with zero wait, zero rejected commits, and the same 17-LOC adapter declaration across effect classes, while every tied baseline pays on at least one of these axes.

\paragraph{Contributions.}
\begin{itemize}[leftmargin=*, nosep]
\item We define a transactional settlement model for agent tool effects that combines sealing, read/effect scopes, per-resource progress predicates, effect-class-aware commit, and shim-based adapter interposition (\S\ref{sec:model}).
\item We implement the model in a runtime that splits tool execution from transaction settlement through adapters, a transaction manager, and a progress tracker (\S\ref{sec:model}; implementation details in Appendix~\ref{app:runtime}).
\item We evaluate fault recovery, frontier-gated isolation, and irreversible-effect gating against recovery and isolation alternatives implemented in a common harness (\S\ref{sec:evaluation}).
\end{itemize}

\paragraph{Scope.}
Atomix is an execution-layer runtime that composes with, rather than replaces, information-flow control, planning-level validation, and kernel-level confinement (\S\ref{sec:related}). We do not claim semantic validation, distributed deployment, or full crash-safe exactly-once. The prototype implements crash-safe deduplication but validates it only with unit tests.

% Section break

\section{Transactional Abstractions and Runtime Semantics}
\label{sec:model}\label{sec:design}\label{sec:background}

Atomix's runtime model separates tool execution from effect settlement. We first define the failure modes, then the runtime abstractions, then the commit and abort rules.

\begin{figure}[htbp]
\centering
\includegraphics[width=\linewidth]{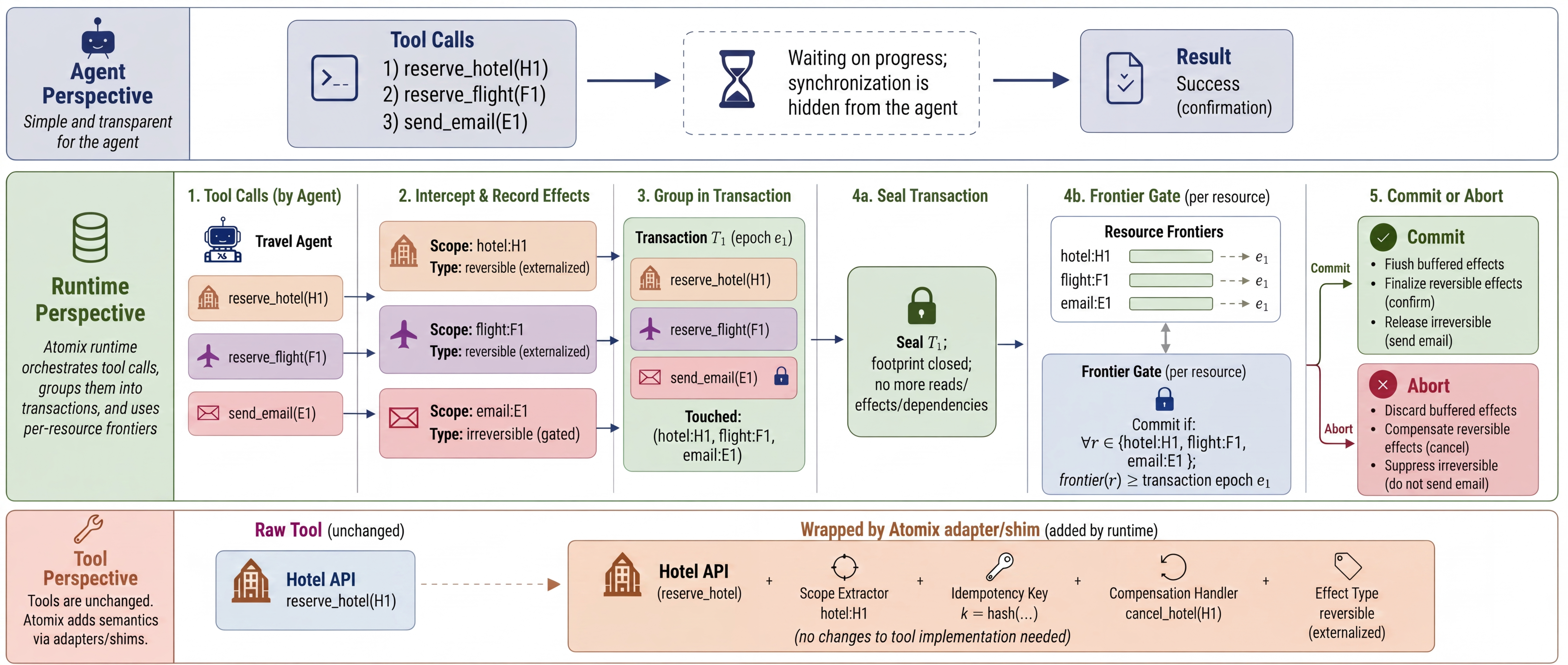}
\figcaption{Atomix lifecycle across agent, runtime, and tool layers. The agent receives synchronous tool returns plus one settlement outcome; the runtime records scopes and effects, seals the transaction footprint, gates settlement on per-resource frontiers, and interposes through tool shims.}
\label{fig:atomix-overview}
\end{figure}

\paragraph{Motivating example.} A holiday-booking agent reserves a flight, a hotel, and a rental car, then sends a confirmation email (Figure~\ref{fig:atomix-overview}). If the flight step fails, the runtime should abort the whole transaction. The four steps have different rollback affordances: the flight hold is refundable, the hotel reservation is compensatable with a cancellation fee, and the confirmation email is irreversible. Reactive compensation cannot gate the email. Under concurrent agents sharing the booking, per-call locks do not span the read--plan--write gap, so stale plans overwrite each other. Under speculative plans, losing branches externalize effects that a snapshot restore cannot roll back. The missing signal is whether the agent work has reached a point where effects can settle safely.

Atomix fills that gap with progress-aware transactions. Inside a transaction, tools run when the agent calls them, but their effects become permanent only when the runtime decides the transaction may settle. Execution, \emph{seal} (freezing the transaction footprint), frontier checking, and commit-or-abort are separate events. Under speculation, each candidate branch follows the same discipline; the controller commits the chosen branch and aborts the others.

\noindent The integration layer chooses transaction granularity: the smallest unit of agent intent whose effects must settle together under one invariant. In practice this is an agent action phase, not a single call or an entire long-running workflow.

\subsection{Core Abstractions}
\label{subsec:model-objects}

Atomix uses five runtime objects:
\begin{itemize}[leftmargin=*, nosep]
\item \textbf{Epochs} are logical timestamps: totally ordered monotonic scalars in the Naiad sense~\cite{murray2013naiad}, assigned by a monotonic allocator. They encode ordering, not identity; a transaction's identifier (\texttt{trace\_id}) is orthogonal.
\item \textbf{Scopes} name resources. A transaction records both read scopes (state the agent observed and depended on) and effect scopes (state the transaction may mutate or release).
\item \textbf{Effects} describe a tool call's side effect: its resource scope, an idempotency key (from the tool or adapter), an effect class, and optional release or compensation handlers.
\item \textbf{Frontiers} are per-resource monotonic cursors on the epoch timeline; the orchestrator advances a resource's frontier once it certifies that no earlier-epoch work on that resource can still arrive.
\item \textbf{Transactions} group the read scopes and effects that must settle together. The integration layer declares the granularity. A transaction can span one tool call, one agent read--plan--write phase, one speculative branch, or a short workflow whose effects share one user-facing invariant. Disjunctive ``any one of $K$'' semantics are not a second transaction shape: each option runs as its own transaction, and the speculation controller commits the chosen branch and aborts the others.
\end{itemize}

\subsection{Lifecycle and Ownership}
\label{subsec:model-lifecycle}

A transaction moves through four states:
\begin{center}
\texttt{open -> sealed -> waiting\_on\_progress -> \{committed, aborted\}}
\end{center}
While a transaction is \texttt{open}, adapters may register read scopes and record effects. The orchestrator \emph{seals} the transaction after every in-flight call under that transaction has returned, recorded its effect, or canceled. Seal is a contract: the transaction may not append new read scopes, effect scopes, effects, or dependencies. Atomix enforces the seal by rejecting late additions and aborting the transaction as a protocol violation. A sealed transaction may still wait on frontiers, fail a pre-commit hook, lose branch selection, or abort because a read became stale.

The ownership split is explicit. The agent chooses tool calls; Atomix does not trust it to perform concurrency control. The orchestrator or integration layer chooses transaction granularity, tracks pending calls, seals transactions, selects speculative winners, aborts losers, and advances frontiers. Adapters name scopes, classify effects, derive idempotency keys, and provide release or compensation handlers. The transaction manager enforces the state machine and effect-class settlement. The progress tracker stores frontiers and decides whether a sealed transaction may commit. Semantic validation, when needed, runs in a pre-commit hook.

The progress tracker implements the commit predicate from \S\ref{subsec:model-transaction-semantics} as per-resource frontiers plus waiters keyed by overlapping scopes. A sealed transaction that is not yet admissible remains queued; a frontier advancement wakes affected waiters and re-evaluates the predicate. Frontier advancement is integration-specific: sequential runs advance after each call completes, DAGs after all predecessors on the scope complete, and speculation only after winner selection and loser finalization. Premature advancement is an orchestrator-contract violation; delayed advancement only stalls settlement.

The transaction manager turns this protocol into runtime behavior. \texttt{begin} opens a transaction at a fresh epoch; \texttt{register\_read} records observed scopes; \texttt{record\_effect} stages the adapter's effect and updates the transaction's commit epoch; \texttt{seal} freezes the footprint; \texttt{request\_commit} asks the progress tracker whether settlement is admissible; and \texttt{abort} settles without releasing gated effects. If a tool produces a new read or effect after seal, the manager rejects the append and aborts the transaction as a protocol violation. Appendix~\ref{app:runtime} gives the full runtime design.

\paragraph{Worked example.}
Figure~\ref{fig:atomix-overview} traces a short booking transaction across the three swim lanes. The reservations (\texttt{reserve\_hotel/flight/car(d1)}) are reversible and execute eagerly, returning the providers' real confirmations. Atomix buffers \texttt{send\_email(d1)} at the adapter gate, gives the agent a synthetic queued-acknowledgment, and fires the call only after per-resource frontiers admit the transaction. \textbf{On commit}, Atomix releases the buffered email, finalizes externalized reservations, and the orchestrator returns \texttt{committed}. \textbf{On abort} (e.g., the flight step fails), compensations cancel the reservations in reverse dependency order and Atomix discards the email. Concurrent rebooking on the same reservation waits for earlier work to finalize. \S\ref{subsec:design-adapters} details the adapter declarations.

\subsection{Effect Taxonomy and Tool Adapters}
\label{subsec:model-effect-taxonomy}
\label{subsec:design-adapters}

Following classical transaction literature~\cite{gray1978notes}, Atomix classifies effects by \emph{reversibility} (whether compensation can undo the effect) and \emph{externalization timing} (whether the tool commits on call or the runtime can defer it). The runtime maps this classification to three execution paths in Algorithm~\ref{alg:lifecycle}. \emph{Reversible} effects (file edits, DB writes; \emph{reversible-with-cost} variants for refundable purchases) execute eagerly and compensate in reverse dependency order on abort. \emph{Bufferable} effects (local filesystem writes, in-memory state) defer execution, return a staged-acknowledgment, and apply on commit or discard on abort. \emph{Irreversible-gated} effects (emails, transfers, physical actions) hold the call at the adapter, return a queued-acknowledgment, and fire only on commit. An optional fourth class, \emph{idempotent-known-outcome} (e.g., payment-intent APIs with server-side idempotency keys), shares the Reversible path without a custom compensation handler. In the travel example, reservations are Reversible-with-cost and confirmation emails are Irreversible-gated.

Tools integrate through adapters rather than rewrites. To expose a tool to Atomix, the adapter supplies a scope extractor, effect builder, idempotency key, effect class, and optional release or compensation handlers. If the exact resource is unknown before execution, the adapter registers a conservative may-touch scope and refines it after the result identifies concrete resources. Coarse scopes fail closed by reducing concurrency; missing or too-narrow scopes fail open by allowing stale reads or conflicting effects. Appendix Table~\ref{tab:adapter-spec} lists representative adapters.

\subsection{Commit and Abort Rules}
\label{subsec:model-transaction-semantics}

The \textbf{commit rule} enforces the progress predicate. For a transaction $T$, let $\mathcal{R}(T)$ be the set of read and effect scopes recorded in its sealed footprint, and let $e(T)$ be the maximum epoch of those recorded reads and effects. A sealed transaction $T$ commits only when $\mathit{frontier}(r) \geq e(T)$ for every $r \in \mathcal{R}(T)$. The orchestrator advances $\mathit{frontier}(r)$ to value $e$ only after every transaction with $\mathit{commit\_epoch} < e$ that touches $r$ has finalized. The strict-less-than makes the predicate non-circular: $T$ at $\mathit{commit\_epoch}\,e$ commits when $\mathit{frontier}(r) \geq e$, depending only on smaller-epoch peers. In Atomix, commit means final settlement: the runtime releases bufferable and irreversible effects at commit and accepts already-externalized reversible effects as final. Externalized reversible effects may be observed before abort, so Atomix relies on compensation rather than the cursor to undo them.

The \textbf{abort rule} replays compensations in reverse dependency order~\cite{dunlap2002revirt}. Abort fires on tool failure, compensation failure, pre-commit veto, losing-branch selection, late post-seal effects, or \emph{stale read} (a waiting transaction whose observed scope an earlier-epoch commit has updated). Atomix tags abort outcomes per effect: \emph{clean} if compensation succeeds; \emph{leaked} if Atomix released any irreversible effect before the abort decision; \emph{unresolved} for compensation failures, residue from reversible-with-cost effects, missing handlers, and \emph{partial-commit} from multiple gated irreversibles where one release succeeded and a later one failed after retry exhaustion (atomic externalization across heterogeneous endpoints is impossible above the tool layer; \S\ref{sec:discussion}). Leaked, unresolved, and partial-commit outcomes surface to the operator through the same residue channel.

The commit path also accepts a \textbf{pre-commit hook}: a user-supplied predicate over the sealed transaction footprint. Atomix uses it as the attachment point for semantic validation, which the runtime leaves out of scope. Algorithm~\ref{alg:frontier} in Appendix~\ref{app:runtime} makes the predicate explicit; shared-resource isolation requires the additional mechanisms in \S\ref{subsec:model-safety}.

\paragraph{Visibility model.} Atomix distinguishes \emph{provider visibility} from \emph{Atomix-mediated visibility}. Reversible-eager effects externalize on call, so the provider (and any observer outside Atomix) sees them before commit; abort runs compensation against that already-visible state. Atomix-mediated visibility is what later transactions see through \texttt{register\_read}: frontiers admit a later $T'$ only after every smaller-epoch transaction on its scopes has finalized, and abort-on-stale retry rolls $T'$ back if an observed version is overwritten pre-commit. The composition prevents \emph{Atomix-mediated} dirty reads and stale-write commits but does not undo external mid-transaction visibility of reversible-eager effects. This is the price of supporting unmodified providers behind the shim. Bufferable and irreversible-gated effects have a single visibility event at commit, so the two regimes coincide.

\subsection{Safety Invariants}
\label{subsec:model-safety}

\paragraph{Trusted computing base.} Three assumptions are load-bearing: (i) adapter mediation of every effect-bearing call; (ii) correct adapter metadata (scopes, idempotency keys, compensations, effect classes); (iii) frontier advancement per the strict-less-than contract. Given these, \emph{atomicity by effect class}: bufferable effects are all-or-nothing, reversible effects compensate on abort (residue only on compensation failure), and irreversibles stay gated (\S\ref{subsec:model-effect-taxonomy}). \emph{Single-process deduplication}: idempotency keys plus durable dedup prevent retry duplicates within one orchestrator process; we do not claim distributed crash-safe exactly-once (Appendix~\ref{app:runtime}). \emph{Progress-safe settlement}: frontier ordering composed with scope-on-read and abort-on-stale retry delivers per-resource commit-order settlement; losing-branch bufferable effects leave no persistent state, irreversibles stay gated, and partitions or crashes stall frontiers.

\paragraph{Implementation.} The Atomix prototype is a ${\sim}$2{,}000-line single-process Python runtime; Appendix~\ref{app:runtime} gives implementation details and limits.

% Section break

\section{Evaluation}
\label{sec:evaluation}

We evaluate five questions: \textbf{RQ1} fault recovery (\S\ref{subsec:rq1}), \textbf{RQ2} frontier-gated isolation under contention and speculation (\S\ref{subsec:rq2}), \textbf{RQ3} irreversible-effect gating (\S\ref{subsec:rq3}), \textbf{RQ4} differentiation from adjacent systems (\S\ref{subsec:capability}), and \textbf{RQ5} runtime cost (Appendix~\ref{app:overhead}). RQ1--RQ3 mirror the three claims of \S\ref{sec:introduction}; Appendices~\ref{app:evaluation-protocol}--\ref{app:related-systems} give the protocol, supporting tables, boundary controls, and the full adjacent-system matrix.

\subsection{Methodology}
\label{subsec:evaluation-methodology}

\paragraph{Workloads, baselines, and statistics.} The primary real-workload fault substrate is $\tau$-bench retail (GPT-4.1, max\_steps=30, $N{=}30$ at $fp\in\{0.10,0.30\}$, $N{=}10$ at $fp{=}0$). WebArena (GPT-5, \texttt{reasoning\_effort=medium}) and OSWorld (Claude Sonnet~4) are secondary real-infrastructure substrates at $fp\in\{0.10,0.30\}$ (Appendix~\ref{app:rq1-results}). Multi-agent $\tau$-bench uses 2/4/8 agents with GPT-4o-mini for real-LLM validation. Controlled tests isolate individual mechanisms: a combined-stress benchmark ($N{=}100$ per cell), a speculation store with $K\in\{2,4,8,16\}$ branches across four effect classes, contention microbenchmarks, and an irreversible-email task. We focus on benchmarks whose tool calls externalize state rather than reasoning- or capability-centric suites such as SWE-bench~\cite{jimenez2024swebench}, GAIA~\cite{mialon2024gaia}, or AgentBench~\cite{liu2024agentbench}. Our harness implements 9 mechanism-matched baselines (Tx-Full, Tx-NoFrontier+Retry, Saga-Compensation, Checkpoint-Replay, Workflow-Lock, Per-Call-Lock, OCC-Revalidate-and-Retry, TCC-Confirm, Mutex+WAL+Rollback, No-Tx), not full system ports; Mutex+WAL+Rollback tracks SAFEFLOW's transactional axis without its information-flow control. Faults inject per-call with Bernoulli probability $fp$ across five classes (F1 pre-execution, F2 post-effect/pre-return, F3 in-compensation, F4 duplicate-delivery, F5 timeout; see Appendix~\ref{app:fault-injection}). The injection rate is a stress knob, not a production base rate. Primary metrics are clean task success (RQ1), conflict-serializability violations (RQ2), and classified-before-externalization leak rate (RQ3); we use 95\% Clopper-Pearson CIs, report zero-counts as 95\% upper bounds, and use Fisher's exact test for significance. Appendix~\ref{app:evaluation-protocol} gives hardware, substrates, and protocol details.

\subsection{RQ1: Fault Recovery on Real LLM-Agent Workloads}
\label{subsec:rq1}

\paragraph{Framing.} RQ1 measures whether a transaction layer that retries faulted calls, compensates externalized reversible effects, and deduplicates ambiguous retries improves end-to-end task success when individual tool calls fail mid-workflow. The clean-success audit checks final state for residue, duplicate effects, and unresolved compensation. The main table compares recovery mechanisms on the same real $\tau$-bench sweep. Appendix~\ref{app:multirate-extended} adds the synthetic cut-point sweep, including Checkpoint-Replay, to localize which fault classes create the separation. RQ1 isolates recovery from isolation: we run mostly sequential workloads and inject faults across the full lifecycle.

\paragraph{Experimental setup.} $N{=}30$ tasks per cell on $\tau$-bench retail across $fp \in \{0, 0.10, 0.30\}$, comparing Tx-Full against seven baselines (parameters in \S\ref{subsec:evaluation-methodology}). Appendix~\ref{app:rq1-results} adds the WebArena and OSWorld sweeps; Table~\ref{tab:f2-f4-split} isolates F2 (post-effect/pre-return) and F4 (duplicate-delivery) faults to localize the mechanism difference.

\begin{figure}[t]
\centering
\begin{tikzpicture}
\begin{axis}[
  width=0.78\linewidth, height=5.6cm, ymin=-8, ymax=108, xmin=-0.45, xmax=2.55,
  xtick={0,1,2}, xticklabels={$fp{=}0$, $fp{=}0.10$, $fp{=}0.30$},
  ylabel={Clean success (\%)}, ylabel near ticks,
  legend style={font=\tiny, at={(1.03,1.0)}, anchor=north west, legend columns=1},
  legend cell align=left, reverse legend, mark size=2.2pt,
  every axis plot/.append style={thick},
  error bars/error bar style={line width=0.5pt}]
\addplot+[mark=diamond*, color=orange!85!black, error bars/.cd, y dir=both, y explicit] table[y error plus=ep, y error minus=em] {
  x y  ep em
   0.28 60 28 34
   1.28 23 19 13
   2.28  7 15  6
}; \addlegendentry{OCC-Reval.+Retry}
\addplot+[mark=x, color=gray, error bars/.cd, y dir=both, y explicit] table[y error plus=ep, y error minus=em] {
  x y  ep em
   0.21 50 31 31
   1.21 27 19 15
   2.21  0 12  0
}; \addlegendentry{No-Tx}
\addplot+[mark=diamond, color=brown!70, error bars/.cd, y dir=both, y explicit] table[y error plus=ep, y error minus=em] {
  x y  ep em
   0.14 60 28 34
   1.14 30 19 15
   2.14  3 14  3
}; \addlegendentry{No-Frontier}
\addplot+[mark=triangle, color=teal!70, error bars/.cd, y dir=both, y explicit] table[y error plus=ep, y error minus=em] {
  x y  ep em
   0.07 60 28 34
   1.07 33 20 16
   2.07  3 14  3
}; \addlegendentry{Mutex+WAL+Rollback}
\addplot+[mark=o, color=violet, error bars/.cd, y dir=both, y explicit] table[y error plus=ep, y error minus=em] {
  x y  ep em
   0.00 60 28 34
   1.00 37 19 17
   2.00  7 15  6
}; \addlegendentry{Saga-Compensation}
\addplot+[mark=square, color=blue!50, error bars/.cd, y dir=both, y explicit] table[y error plus=ep, y error minus=em] {
  x y  ep em
  -0.07 60 28 34
   0.93 40 19 17
   1.93  3 14  3
}; \addlegendentry{TCC-Confirm}
\addplot+[mark=square*, color=blue!70!black, line width=1.3pt, error bars/.cd, y dir=both, y explicit] table[y error plus=ep, y error minus=em] {
  x y  ep em
  -0.14 60 28 34
   0.86 63 17 19
   1.86 53 19 19
}; \addlegendentry{\textbf{Checkpoint-Replay}}
\addplot+[mark=*, color=red!80!black, line width=1.5pt, error bars/.cd, y dir=both, y explicit] table[y error plus=ep, y error minus=em] {
  x y  ep em
  -0.21 60 28 34
   0.79 73 14 19
   1.79 57 18 20
}; \addlegendentry{\textbf{Tx-Full}}
\end{axis}
\end{tikzpicture}
\caption{Clean task success on $\tau$-bench retail (GPT-4.1, max\_steps=30) across fault rates. Bars show 95\% Clopper-Pearson CIs; faulted cells use $N{=}30$ and the $fp{=}0$ calibration uses $N{=}10$.}
\label{fig:rq1-multirate}\label{tab:rq1-multirate-main}
\end{figure}

\paragraph{Analysis.} Figure~\ref{fig:rq1-multirate} gives the real $\tau$-bench fault sweep. At $fp{=}0.30$, Tx-Full reaches 57\% clean task success, while TCC, Saga, Mutex+WAL+Rollback, No-Frontier, No-Tx, and OCC fall to 0--7\%. Checkpoint-Replay is statistically tied on RQ1 (53\%, 16/30 vs.\ Tx-Full's 17/30, Fisher $p{=}1.0$; the tie also holds at the full $N{=}114$ task pool, $p{\approx}0.50$, Appendix~\ref{app:rq1-results}); RQ3 and combined stress distinguish the two mechanisms. The F2/F4 isolation run in Table~\ref{tab:f2-f4-split} shows that the recovery separation is not a duplicate-delivery artifact: idempotent retry handles F4, while F2 post-effect/pre-return failures are where Atomix's transactional settlement helps most.

\subsection{RQ2: Frontier-Gated Isolation}
\label{subsec:rq2}

\paragraph{Framing and setup.} Stale-plan races in concurrent agents pass through per-call locks (released between read and write) but force workflow-scoped locks to serialize disjoint work. Atomix's claim is that frontier ordering plus scope-on-read plus abort-on-stale retry matches workflow-lock safety without that cost. The multi-agent $\tau$-bench harness instantiates the booking race on retail orders with two regimes (\emph{forced-overlap}: same order with cancel-vs.-address-modify intents; \emph{disjoint}: different orders), comparing Tx-Full against Workflow-Lock, Per-Call-Lock, OCC, and No-Tx. The synthetic plan-delay variant runs 100 trials/cell at 4 agents forced-overlap and 2/4/8 agents disjoint; the real-LLM variant inserts a GPT-4o-mini call between phases (100 trials at 4 ag, 30 at 8 ag). Appendix~\ref{app:rq2-results} reports speculation isolation on a controlled 4-effect-class benchmark ($K\in\{2,4,8,16\}$).

\begin{table}[t]
\centering
\footnotesize
\setlength{\tabcolsep}{3pt}
\begin{tabular}{@{}lccccccc@{}}
\toprule
& \multicolumn{3}{c}{Synthetic plan-delay (4 ag, $N{=}100$)} & \multicolumn{2}{c}{Real LLM (GPT-4o-mini)} & \multicolumn{2}{c}{Disjoint wait (ms)} \\
\cmidrule(r){2-4}\cmidrule(lr){5-6}\cmidrule(l){7-8}
Mode & Inv./run & Wait (ms) & Blocked & Inv./run, 4ag & Inv./run, 8ag & 4 ag & 8 ag \\
\midrule
\textbf{Tx-Full}          & \textbf{0.00}$^{\dagger}$ & \phantom{1}\textbf{0.0} & \textbf{0\%} & \textbf{0.0}$^{\dagger}$ & \textbf{0.0} & \textbf{0.0} & \textbf{0.0} \\
Workflow-Lock             & 0.00$^{\dagger}$ & 112.3 & 75\% & 0.0$^{\dagger}$ & 0.0 & 44.9 & 104.7 \\
Per-Call-Lock             & 4.91 & \phantom{1}0.0 & 0\% & 3.0 & 7.0 & 0.0 & 0.0 \\
OCC-Revalidate-and-Retry  & 0.00$^{\ddagger}$ & \phantom{1}0.0 & 0\% & 0.0$^{\ddagger}$ & 0.0$^{\ddagger}$ & 0.0 & 0.0 \\
No-Tx                     & 4.97 & \phantom{1}0.0 & 0\% & 3.0 & 7.0 & 0.0 & 0.0 \\
\bottomrule
\end{tabular}
\vspace{1mm}
\begin{minipage}{0.96\linewidth}
\footnotesize $^{\dagger}$0 invariant violations and 0 conflict-cycle witnesses across 100 schedule traces (CP-95\% upper bound $\leq$3.7\%). Here, the invariant fires precisely when two committed writes form a conflict cycle on the same scope (cancel commits read-then-writes a state that a concurrent transaction's read-then-write had already modified), so invariant violations and conflict-cycle witnesses are equal by construction; the upper bound reported here applies to both metrics. $^{\ddagger}$OCC's zero-invariant rows come from rejecting stale writes rather than committing all work; the per-run rejection rate equals the conflict rate (3.00 OCC rejections/run at 4 agents, 7.00 at 8 agents in the real-LLM substrate, matching the 3.0 and 7.0 invariant violations per run that Per-Call-Lock and No-Tx commit at the same agent counts).
\end{minipage}
\caption{Multi-agent $\tau$-bench contention results. \emph{Inv./run} is the workload invariant diagnostic; an independent operation log checks conflict-serializability. Wait columns report average blocking time.}
\label{tab:rq2-multiagent}
\end{table}

\paragraph{Analysis.} Table~\ref{tab:rq2-multiagent} separates safety, wait, and disjoint-resource throughput. Tx-Full and OCC are the two safety-clean modes with zero wait. Workflow-Lock matches on safety but pays 112~ms wait under contention and 45/105~ms even on disjoint resources. Per-Call-Lock and No-Tx commit 4.9 invariants/run (each one a conflict-cycle witness). OCC sidesteps via rejection at the cost of 3{,}000 rejected commits per 100 runs at 4 agents. The real-LLM 8-agent block reproduces the qualitative pattern at a higher per-run rate; Tx-Full holds 0 across all configurations. The serializability checker reports no conflict-cycle witnesses for Tx-Full or Workflow-Lock.

\paragraph{Composition ablation.} An ablation removing each of scope-on-read, abort-on-stale, or per-resource frontiers in turn shows that all three are load-bearing: removing either of the first two collapses to Mutex-Per-Resource (5 invariants/run); removing the third is safety-clean but pays 113~ms wait (Appendix~\ref{app:rq2-results}, Table~\ref{tab:e2-ablations}).

\paragraph{Wait scaling.} A global-lock baseline (Tx-GlobalFrontier) pays approximately linear-in-$n$ wait per task at higher agent counts (564~ms at $n{=}16$, 2{,}369~ms at $n{=}64$); Tx-Full holds 0~ms wait at every $n$ via per-resource granularity (Appendix~\ref{app:rq2-results}, Table~\ref{tab:wait-scaling}).

\paragraph{Result.} In these trials, the contention checker finds no conflict-cycle witnesses for Tx-Full; per-resource granularity avoids Workflow-Lock's wait on disjoint resources; and the speculation benchmark reports no persistent losing-branch residue when irreversible effects remain gated.

\subsection{RQ3: Irreversible-Effect Gating}
\label{subsec:rq3}

\paragraph{Framing.} Effects that cannot be undone once externalized (emails, payments, deletions) fall outside the reach of compensation: a Saga can refund a charge but cannot un-send an email. Gated commit prevents leakage by holding the irreversible effect at the adapter gate until the commit predicate fires. RQ3 asks whether gated commit prevents leakage under enumerated abort sources, and whether effect-class classification costs less developer effort than per-tool TCC handlers or per-resource Mutex+WAL+Rollback gating.

\paragraph{Experimental setup.} A mixed reversible/irreversible workflow uses a real SMTP/webhook sink, 100 trials per (abort source $\times$ baseline) cell across 5 abort sources (tool failure, losing speculation, stale read, pre-commit veto, timeout), and 7 baselines. Each row has 500 invalid-send trials and 500 valid-send positive-control trials. The leak metric counts effects classified before externalization that become visible; the harness logs post-externalization residue separately when the provider commits before the adapter records the call (Appendix~\ref{app:compfail}). Every baseline delivers 500/500 valid sends.

\begin{table}[t]
\centering
\footnotesize
\setlength{\tabcolsep}{3pt}
\begin{tabular}{@{}lcccc@{}}
\toprule
Mode & Leaks & Attempts & Leak rate & 95\% CI \\
\midrule
\textbf{Tx-Full} & \textbf{0} & 500 & \textbf{0\%} & \textbf{[0, 0.74]} \\
Saga-Compensation & 400 & 500 & 80\% & [76, 83] \\
Checkpoint-Replay & 200 & 500 & 40\% & [36, 44] \\
TCC-Confirm & 0 & 500 & 0\% & [0, 0.74] \\
Mutex+WAL+Rollback$^{\ast}$ & 0 & 500 & 0\% & [0, 0.74] \\
No-Tx & 500 & 500 & 100\% & [99.3, 100] \\
Atomix-MisclassifiedIrreversible & 300 & 500 & 60\% & [56, 64] \\
\bottomrule
\end{tabular}
\caption[Irreversible-effect leakage against an append-only sink]{Irreversible-effect leakage against a real append-only sink. Each row sums 5 abort sources with $N{=}100$ trials each, for 500 invalid-send trials per baseline. The final column gives the Clopper-Pearson interval for leak rate.}
\label{tab:rq3-irrev}\label{tab:e4-irrev}
\end{table}

\paragraph{Analysis.} Table~\ref{tab:rq3-irrev} gives invalid-send leakage against the append-only sink. The paired valid-send positive control passes at 500/500 across every baseline, so the gate does not suppress commits that should be released. Three mechanisms hold zero classified-before-externalization leaks: Tx-Full, TCC-Confirm, and Mutex+WAL+Rollback. Saga-Compensation leaks on 4 of 5 abort sources (80\%) because compensation cannot un-send; Checkpoint-Replay leaks under tool-failure and timeout (40\%) because retried calls re-externalize; the misclassification ablation leaks 60\% because the gate never fires. TCC and Mutex+WAL+Rollback match Atomix at zero leaks only because they are fully wired for the email tool here. Under default DB-only wiring they leak the mailbox class on every losing branch (Appendix~\ref{app:rq2-results}). The differentiation is integration cost: each requires per-tool wiring (50 and 30 LOC respectively, scaling linearly with tool count) versus Atomix's 17 LOC/adapter regardless of effect class (Appendix~\ref{app:rq3-results}, Table~\ref{tab:rq3-cost}).

\subsection{Combined Stress}
\label{subsec:joint-surface}

\paragraph{Framing.} RQ1--RQ3 each isolate one failure mode. The whole-system claim is the joint case: clean recovery, no irreversible leaks, no stale writes, no losing-branch residue, and no global serialization. Different baselines achieve one axis (TCC and Mutex+WAL+Rollback gate irreversibles; Workflow-Lock and OCC isolate stale plans; Saga and Checkpoint-Replay support recovery), but none covers the union under contention, faults, and irreversible confirmation.

\paragraph{Setup.} A combined-stress synthetic benchmark exercises all five axes: 4 agents, 2 contended orders (cancel/modify pairs), per-task workflow read $\to$ plan-pause $\to$ write $\to$ irreversible confirmation. Faults inject between write and confirmation with $f_\text{class} \in \{F2, F4, \text{mixed}\}$ at $fp \in \{0.10, 0.30\}$. Each run records per-agent clean outcome, invariant violations (cancelled-and-modified on the same order), stale commits, confirmation leaks, duplicates, and wait time. Table~\ref{tab:joint-surface} defines \emph{run-clean}.

\begin{table}[H]
\centering
\footnotesize
\setlength{\tabcolsep}{3pt}
\begin{tabular}{@{}lccccc@{}}
\toprule
& \multicolumn{2}{c}{Mixed faults} & F2-only & F4-only & Wait \\
\cmidrule(r){2-3}
Mode & $fp{=}0.10$ & $fp{=}0.30$ & $fp{=}0.10$ & $fp{=}0.10$ & (ms) \\
\midrule
\textbf{Tx-Full}      & \textbf{84} & \textbf{65} & \textbf{100} & 85 & \phantom{1}0.1 \\
OCC                   & 82 & 62 & \textbf{100} & 75 & \phantom{1}0.0 \\
Mutex-Workflow        & 81 & 62 & \textbf{100} & 77 & 95.9 \\
TCC-Confirm           & 79 & 63 & \textbf{100} & 81 & \phantom{1}0.1 \\
Mutex+WAL+Rollback    & 70 & 24 & 80 & 78 & \phantom{1}0.4 \\
Saga-Compensation     & 68 & 24 & 79 & 87 & \phantom{1}0.1 \\
No-Tx                 & 67 & 17 & 86 & 86 & \phantom{1}0.0 \\
Tx-NoAbortOnStale     & 65 & 25 & 83 & 79 & \phantom{1}0.1 \\
Mutex-Per-Resource    & 64 & 23 & 85 & 83 & \phantom{1}0.1 \\
Checkpoint-Replay     & 63 & 25 & 81 & 74 & \phantom{1}0.1 \\
\bottomrule
\end{tabular}
\caption{Run-clean rate on combined-stress ($N{=}300$ mixed, $N{=}100$ F2/F4-only, 4 agents, 2 contended orders). Run-clean requires zero invariant violations, stale commits, confirmation leaks, and duplicates. Wait is from mixed $fp{=}0.10$.}
\label{tab:joint-surface}
\end{table}

\paragraph{Analysis.} Tx-Full is Pareto-best on the joint surface. Four mechanisms tie on run-clean rate across both fp regimes (Tx-Full, OCC, Mutex-Workflow, TCC-Confirm; Fisher's $p>0.1$ for every Tx-Full comparison), but each tied baseline pays on a secondary axis: Mutex-Workflow waits 95.9~ms, OCC rejects commits (\S\ref{subsec:rq2}), TCC requires per-tool wiring (Appendix~\ref{app:rq3-results}); Tx-Full alone reports 0~ms wait, 0 rejections, and a 17-LOC adapter declaration across effect classes. Tx-NoAbortOnStale at 25--65\% is the largest single-primitive ablation drop.

\paragraph{Real-LLM validation.} Replacing the synthetic plan-pause with a real GPT-4o-mini call between read and write phases reproduces the tier structure (Tx-Full 90\%, TCC 93\%, Saga 80\%, CR 73\% run-clean at $fp{=}0.10$, $N{=}30$); the wait gap against Mutex-Workflow widens because the global lock blocks every concurrent agent for the full ${\sim}$700\,ms LLM call window (Appendix~\ref{app:rq2-results}).

\paragraph{Capability comparison.} \label{subsec:capability} RQ1--RQ3 each show one classical mechanism matching Atomix on one safety axis (Workflow-Lock on contention; TCC and Mutex+WAL+Rollback on irreversible-effect leakage). Appendix~\ref{app:related-systems} (Table~\ref{tab:capability}) gives the union comparison: no compared system covers all seven axes (progress predicates, effect classes, speculation isolation, irreversible gating, stale-plan isolation, shim deployment, single-process deduplication) under shim-only deployment. SAFEFLOW is closest on safety but assumes a co-designed stack with information-flow control; durable-execution engines and watermark systems each cover a subset.

\section{Discussion and Limitations}
\label{sec:discussion}

\paragraph{Enforcement boundary and metadata.} Atomix mediates only calls that enter its adapters; bypass paths lie outside the enforcement boundary. Adapter metadata (scopes, idempotency keys, compensation handlers, effect classes) is safety-critical: too-narrow scopes commit 66/200 invariant violations and a wrong effect class leaks 17/200 irreversibles (Appendix~\ref{app:annotations}). The released aliasing test suite (Appendix~\ref{app:aliasing}) partially validates this surface; a linter and scope fuzzer remain future work. The orchestrator must also honor the frontier-advancement contract (Appendix~\ref{app:e5-b2}).

\paragraph{Out of scope.} Semantic validation, distributed deployment, and full crash-safe exactly-once are not runtime contributions; the prototype validates single-process deduplication only. Information-flow control and planning-layer validation are orthogonal and compose with Atomix.

\paragraph{Distributed deployment sketch.} The model appears compatible with distributed implementations: per-resource cursors with monotonic advancement resemble Naiad's pointstamps~\cite{murray2013naiad} and Flink/Dataflow watermarks~\cite{flinkdocs,dataflow_exactlyonce}. A distributed Atomix would require a sharded frontier store with consensus advancement (Calvin-style~\cite{thomson2012calvin}), a replicated effect log for crash-safe exactly-once (Raft, applied to the effect record), and adapter-side idempotency keys preserved across replica failover.

\paragraph{Atomic externalization across multiple irreversibles.} A transaction with $\geq 2$ gated irreversibles releases them sequentially; a persistent failure after the first has externalized leaves a partial-commit state. Above the tool layer, the runtime has no way to avoid this: atomic commit across heterogeneous external endpoints requires tool-side TCC or 2PC, which arbitrary REST APIs, SMTP servers, and filesystem syscalls do not provide. Saga, Temporal, SAFEFLOW, and OCC hit the same boundary. Only tool-side commit protocols remove it. Atomix bounds the blast radius via bounded retry, fail-stop on persistent failure, and a partial-commit record naming what externalized and what failed.

\paragraph{Responsible deployment and liveness.} Atomix reduces accidental leaks but makes adapter mediation and permission boundaries part of the safety case; deployments should combine it with least-privilege tool access, audit logging, and policy checks for irreversible actions. Hot resources serialize by design. Hung agents can stall frontier advancement until the orchestrator times out. Both are deployment concerns rather than runtime bugs.

\section{Related Work}
\label{sec:related}

\paragraph{Adjacent systems.} Workflow engines and durable-execution platforms (Temporal~\cite{temporal_docs}, AWS Step Functions~\cite{aws_step_functions}, LangGraph~\cite{langgraph_docs}, CrewAI~\cite{crewai_docs}, AutoGen~\cite{wu2023autogen}, Inngest~\cite{inngest_docs}, Restate~\cite{restate_docs}, Cloudflare Workflows~\cite{cloudflare_workflows_docs}, DBOS~\cite{dbos2022}) supply durable execution and retries but do not gate external effect settlement on per-resource workflow progress for unmodified tools. The transactional outbox pattern~\cite{richardson_outbox} is the closest production analog for irreversible-effect gating. Atomix generalizes it to non-database effects (emails, webhooks, physical actions) with adapter-defined keys, plus per-resource progress so multiple agents settle without serializing on a global outbox. SAFEFLOW~\cite{safeflow2025} is the closest safety comparator but assumes a co-designed stack with information-flow control. AgentGit~\cite{agentgit2025}, GoEX~\cite{patil2024goex}, and Speculative Actions~\cite{speculativeactions2024} version state, argue for undo, and predict actions; none supply Atomix's cross-call progress signal. SagaLLM~\cite{chang2025sagallm}, ALAS~\cite{geng2025alas}, MAS-FIRE~\cite{masfire2026}, PEARL~\cite{pearl2026}, Sherlock~\cite{sherlock2025}, and AsyncLM~\cite{asynclm2024} address planning-layer validation, async calls, and agent-induced fault characterization; they compose with Atomix through the pre-commit hook.

\paragraph{Novelty.} Atomix composes Sagas~\cite{garcia1987sagas}, Try-Confirm-Cancel~\cite{pardon2014tcc}, write-ahead logging~\cite{mohan1992aries}, escrow locks~\cite{oneil1986escrow}, deterministic scheduling~\cite{thomson2012calvin}, and streaming watermarks~\cite{murray2013naiad,begoli2021watermarks,flinkdocs,dataflow_exactlyonce,chandy1985snapshot} into a four-event split (execute, seal, frontier-check, settle) at the tool boundary. Each event depends on a different ingredient: adapters classify effects, the transaction manager seals footprints, the progress tracker advances per-resource frontiers, and effect-class settlement chooses between flush, gate-fire, and accept-as-final. In our comparison, each classical primitive omits or conflates at least one ingredient at the tool boundary (Appendix~\ref{app:novelty-matrix}, Table~\ref{tab:novelty-matrix}; full matrix in Table~\ref{tab:capability}).

\section{Conclusion}
\label{sec:conclusion}

LLM agents mutate external state, but current orchestrators cannot transactionally settle tool effects under faults, speculation, contention, and irreversibility. Atomix adds sealed footprints, per-resource progress, scope-on-read with abort-on-stale retry, and effect-class-aware settlement behind shim adapters; the composition closes a gap each prior mechanism covers only partially. Empirically, Atomix is Pareto-best on the joint surface: top run-clean tier at zero wait, zero rejections, and 17 LOC per adapter. Distributed deployment remains future work.

\bibliographystyle{plainnat}
\bibliography{main}

@inproceedings{murray2013naiad,
  title        = {{Naiad}: A Timely Dataflow System},
  author       = {Murray, Derek Gordon and McSherry, Frank and Isaacs, Rebecca and Isard, Michael and Barham, Paul and Abadi, Mart{\'\i}n},
  booktitle    = {Proceedings of the Twenty-Fourth ACM Symposium on Operating Systems Principles (SOSP '13)},
  pages        = {439--455},
  publisher    = {{ACM}},
  year         = {2013},
  doi          = {10.1145/2517349.2522738}
}

@article{chang2025sagallm,
  title        = {{SagaLLM}: Context Management, Validation, and Transaction Guarantees for Multi-Agent {LLM} Planning},
  author       = {Chang, Edward Y. and Geng, Longling},
  journal      = {Proceedings of the {VLDB} Endowment},
  volume       = {18},
  number       = {12},
  pages        = {4874--4886},
  year         = {2025},
  doi          = {10.14778/3750601.3750611}
}

@article{geng2025alas,
  title        = {{ALAS}: Transactional and Dynamic Multi-Agent {LLM} Planning},
  author       = {Geng, Longling and Chang, Edward Y.},
  journal      = {arXiv preprint arXiv:2511.03094},
  year         = {2025},
  doi          = {10.48550/arXiv.2511.03094},
  url          = {https://arxiv.org/abs/2511.03094}
}

@article{safeflow2025,
  title        = {{SAFEFLOW}: A Principled Protocol for Trustworthy and Transactional Autonomous Agent Systems},
  author       = {Li, Peiran and Zou, Xinkai and Wu, Zhuohang and Li, Ruifeng and Xing, Shuo and Zheng, Hanwen and Hu, Zhikai and Wang, Yuping and Li, Haoxi and Yuan, Qin and Zhang, Yingmo and Tu, Zhengzhong},
  journal      = {arXiv preprint arXiv:2506.07564},
  year         = {2025},
  doi          = {10.48550/arXiv.2506.07564},
  url          = {https://arxiv.org/abs/2506.07564}
}

@article{masfire2026,
  title        = {{MAS-FIRE}: Fault Injection and Reliability Evaluation for {LLM}-Based Multi-Agent Systems},
  author       = {Jia, Jin and Deng, Zhiling and Chen, Zhuangbin and Wang, Yingqi and Zheng, Zibin},
  journal      = {arXiv preprint arXiv:2602.19843},
  year         = {2026},
  doi          = {10.48550/arXiv.2602.19843},
  url          = {https://arxiv.org/abs/2602.19843}
}

@inproceedings{pearl2026,
  title        = {{PEARL}: Plan Exploration and Adaptive Reinforcement Learning for Multihop Tool Use},
  author       = {Wang, Qihao and Lu, Mingzhe and Wu, Jiayue and Hu, Yue and Liu, Yanbing},
  editor       = {Mei, Yi and Qian, Chao and Bai, Quan and Xue, Bing and Khanna, Sankalp},
  booktitle    = {{PRICAI} 2025: Trends in Artificial Intelligence -- 22nd Pacific Rim International Conference on Artificial Intelligence, Wellington, New Zealand, November 17--21, 2025, Proceedings, Part {IV}},
  series       = {Lecture Notes in Computer Science},
  volume       = {16454},
  pages        = {639--651},
  publisher    = {Springer, Singapore},
  year         = {2026},
  address      = {Wellington, New Zealand},
  doi          = {10.1007/978-981-95-7081-2_44}
}

@inproceedings{wu2023autogen,
  title        = {{AutoGen}: Enabling Next-Gen {LLM} Applications via Multi-Agent Conversations},
  author       = {Wu, Qingyun and Bansal, Gagan and Zhang, Jieyu and Wu, Yiran and Li, Beibin and Zhu, Erkang and Jiang, Li and Zhang, Xiaoyun and Zhang, Shaokun and Liu, Jiale and Awadallah, Ahmed Hassan and White, Ryen W. and Burger, Doug and Wang, Chi},
  booktitle    = {First Conference on Language Modeling (COLM 2024)},
  publisher    = {OpenReview.net},
  year         = {2024},
  url          = {https://openreview.net/forum?id=BAakY1hNKS}
}

@inproceedings{speculativeactions2024,
  title        = {Speculative Actions: A Lossless Framework for Faster {AI} Agents},
  author       = {Ye, Naimeng and Ahuja, Arnav and Liargkovas, Georgios and Lu, Yunan and Kaffes, Kostis and Peng, Tianyi},
  booktitle    = {The Fourteenth International Conference on Learning Representations (ICLR 2026)},
  publisher    = {OpenReview.net},
  year         = {2026},
  note         = {Oral presentation},
  url          = {https://openreview.net/forum?id=P0GOk5wslg}
}

@article{agentgit2025,
  title        = {{AgentGit}: A Version Control Framework for Reliable and Scalable {LLM}-Powered Multi-Agent Systems},
  author       = {Li, Yang and Ping, Siqi and Chen, Xiyu and Qi, Xiaojian and Wang, Zigan and Luo, Ye and Zhang, Xiaowei},
  journal      = {arXiv preprint arXiv:2511.00628},
  year         = {2025},
  doi          = {10.48550/arXiv.2511.00628},
  url          = {https://arxiv.org/abs/2511.00628}
}

@misc{langgraph_docs,
  title        = {{LangGraph} overview},
  author       = {{LangChain Inc.}},
  howpublished = {\url{https://docs.langchain.com/oss/python/langgraph/overview}},
  year         = {2026},
  note         = {Accessed 2026-05-07}
}

@misc{temporal_docs,
  title        = {{Temporal Docs} | {Temporal Platform Documentation}},
  author       = {{Temporal Technologies Inc.}},
  howpublished = {\url{https://docs.temporal.io/}},
  year         = {2026},
  note         = {Accessed 2026-05-07}
}

@misc{aws_step_functions,
  title        = {What is {Step Functions}?},
  author       = {{Amazon Web Services}},
  howpublished = {\url{https://docs.aws.amazon.com/step-functions/latest/dg/welcome.html}},
  year         = {2026},
  note         = {Accessed 2026-05-07}
}

@misc{crewai_docs,
  title        = {{CrewAI} Documentation},
  author       = {{CrewAI, Inc.}},
  howpublished = {\url{https://docs.crewai.com/}},
  year         = {2026},
  note         = {Accessed 2026-05-07}
}

@misc{mcp_spec_2025,
  title        = {Model Context Protocol Specification, Version 2025-11-25},
  author       = {{Model Context Protocol Project}},
  howpublished = {\url{https://modelcontextprotocol.io/specification/2025-11-25}},
  year         = {2025},
  note         = {Accessed 2026-05-07}
}

@misc{openai_function_calling_guide,
  title        = {Function calling | {OpenAI} {API}},
  author       = {{OpenAI}},
  howpublished = {\url{https://developers.openai.com/api/docs/guides/function-calling}},
  year         = {2026},
  note         = {Accessed 2026-05-07}
}

@misc{claude_code_docs,
  title        = {{Claude Code} overview},
  author       = {{Anthropic}},
  howpublished = {\url{https://code.claude.com/docs/en/overview}},
  year         = {2026},
  note         = {Accessed 2026-05-07}
}

@incollection{gray1978notes,
  title        = {Notes on Data Base Operating Systems},
  author       = {Gray, J. N.},
  editor       = {Bayer, R. and Graham, R. M. and Seegm{\"u}ller, G.},
  booktitle    = {Operating Systems: An Advanced Course},
  series       = {Lecture Notes in Computer Science},
  volume       = {60},
  pages        = {393--481},
  publisher    = {Springer, Berlin, Heidelberg},
  year         = {1978},
  isbn         = {978-3-540-08755-7},
  doi          = {10.1007/3-540-08755-9_9}
}

@article{mohan1992aries,
  title        = {{ARIES}: A Transaction Recovery Method Supporting Fine-Granularity Locking and Partial Rollbacks Using Write-Ahead Logging},
  author       = {Mohan, C. and Haderle, Don and Lindsay, Bruce G. and Pirahesh, Hamid and Schwarz, Peter M.},
  journal      = {{ACM} Transactions on Database Systems},
  volume       = {17},
  number       = {1},
  pages        = {94--162},
  year         = {1992},
  doi          = {10.1145/128765.128770}
}

@inproceedings{dunlap2002revirt,
  title        = {{ReVirt}: Enabling Intrusion Analysis Through Virtual-Machine Logging and Replay},
  author       = {Dunlap, George W. and King, Samuel T. and Cinar, Sukru and Basrai, Murtaza A. and Chen, Peter M.},
  booktitle    = {Proceedings of the 5th Symposium on Operating Systems Design and Implementation (OSDI '02)},
  pages        = {211--224},
  year         = {2002},
  address      = {Boston, MA},
  publisher    = {USENIX Association},
  doi          = {10.1145/1060289.1060309}
}

@inproceedings{garcia1987sagas,
  title        = {Sagas},
  author       = {Garcia-Molina, Hector and Salem, Kenneth},
  booktitle    = {Proceedings of the 1987 {ACM} {SIGMOD} International Conference on Management of Data (SIGMOD '87)},
  pages        = {249--259},
  publisher    = {{ACM}},
  year         = {1987},
  doi          = {10.1145/38713.38742}
}

@misc{flinkdocs,
  title        = {Stateful Stream Processing},
  author       = {{Apache Flink}},
  howpublished = {\url{https://nightlies.apache.org/flink/flink-docs-release-2.2/docs/concepts/stateful-stream-processing/}},
  year         = {2026},
  note         = {Accessed 2026-05-07}
}

@misc{dataflow_exactlyonce,
  title        = {Exactly-once in {Dataflow}},
  author       = {{Google Cloud}},
  howpublished = {\url{https://cloud.google.com/dataflow/docs/concepts/exactly-once}},
  year         = {2026},
  note         = {Accessed 2026-05-07}
}

@article{begoli2021watermarks,
  title        = {Watermarks in Stream Processing Systems: Semantics and Comparative Analysis of {Apache Flink} and {Google Cloud Dataflow}},
  author       = {Akidau, Tyler and Begoli, Edmon and Chernyak, Slava and Hueske, Fabian and Knight, Kathryn and Knowles, Kenneth and Mills, Daniel and Sotolongo, Dan},
  journal      = {Proceedings of the {VLDB} Endowment},
  volume       = {14},
  number       = {12},
  pages        = {3135--3147},
  year         = {2021},
  doi          = {10.14778/3476311.3476389}
}

@article{chandy1985snapshot,
  title        = {Distributed Snapshots: Determining Global States of Distributed Systems},
  author       = {Chandy, K. Mani and Lamport, Leslie},
  journal      = {{ACM} Transactions on Computer Systems},
  volume       = {3},
  number       = {1},
  pages        = {63--75},
  year         = {1985},
  doi          = {10.1145/214451.214456}
}

@article{asynclm2024,
  title        = {Asynchronous {LLM} Function Calling},
  author       = {Gim, In and Lee, Seung-seob and Zhong, Lin},
  journal      = {arXiv preprint arXiv:2412.07017},
  year         = {2024},
  doi          = {10.48550/arXiv.2412.07017},
  url          = {https://arxiv.org/abs/2412.07017}
}

@article{sherlock2025,
  title        = {Sherlock: Reliable and Efficient Agentic Workflow Execution},
  author       = {Ro, Yeonju and Qiu, Haoran and Goiri, {\'I}{\~n}igo and Fonseca, Rodrigo and Bianchini, Ricardo and Akella, Aditya and Wang, Zhangyang and Erez, Mattan and Choukse, Esha},
  journal      = {arXiv preprint arXiv:2511.00330},
  year         = {2025},
  doi          = {10.48550/arXiv.2511.00330},
  url          = {https://arxiv.org/abs/2511.00330}
}

@article{oneil1986escrow,
  title        = {The Escrow Transactional Method},
  author       = {O'Neil, Patrick E.},
  journal      = {{ACM} Transactions on Database Systems},
  volume       = {11},
  number       = {4},
  pages        = {405--430},
  year         = {1986},
  doi          = {10.1145/7239.7265}
}

@inproceedings{thomson2012calvin,
  title        = {Calvin: Fast Distributed Transactions for Partitioned Database Systems},
  author       = {Thomson, Alexander and Diamond, Thaddeus and Weng, Shu-Chun and Ren, Kun and Shao, Philip and Abadi, Daniel J.},
  booktitle    = {Proceedings of the 2012 {ACM} {SIGMOD} International Conference on Management of Data (SIGMOD '12)},
  pages        = {1--12},
  publisher    = {{ACM}},
  year         = {2012},
  doi          = {10.1145/2213836.2213838}
}

@inproceedings{pardon2014tcc,
  title        = {Atomic Distributed Transactions: A {RESTful} Design},
  author       = {Pardon, Guy and Pautasso, Cesare},
  booktitle    = {Proceedings of the 23rd International Conference on World Wide Web (WWW '14 Companion), 5th International Workshop on {Web} {APIs} and {RESTful} Design (WS-REST 2014)},
  pages        = {943--948},
  publisher    = {{ACM}},
  year         = {2014},
  doi          = {10.1145/2567948.2579221}
}

@article{patil2024goex,
  title        = {{GoEX}: Perspectives and Designs Towards a Runtime for Autonomous {LLM} Applications},
  author       = {Patil, Shishir G. and Zhang, Tianjun and Fang, Vivian and {Noppapon C.} and Huang, Roy and Hao, Aaron and Casado, Martin and Gonzalez, Joseph E. and Popa, Raluca Ada and Stoica, Ion},
  journal      = {arXiv preprint arXiv:2404.06921},
  year         = {2024},
  doi          = {10.48550/arXiv.2404.06921},
  url          = {https://arxiv.org/abs/2404.06921}
}

@article{weikum1991principles,
  title        = {Principles and Realization Strategies of Multilevel Transaction Management},
  author       = {Weikum, Gerhard},
  journal      = {{ACM} Transactions on Database Systems},
  volume       = {16},
  number       = {1},
  pages        = {132--180},
  year         = {1991},
  doi          = {10.1145/103140.103145}
}

@misc{inngest_docs,
  title        = {How {Inngest} functions are executed: Durable Execution},
  author       = {{Inngest Inc.}},
  year         = {2025},
  url          = {https://www.inngest.com/docs/learn/how-functions-are-executed},
  note         = {Accessed 2026-05-07}
}

@misc{restate_docs,
  title        = {Welcome to {Restate}!},
  author       = {{Restate Software, Inc.}},
  year         = {2026},
  url          = {https://docs.restate.dev/},
  note         = {Accessed 2026-05-07}
}

@misc{cloudflare_workflows_docs,
  title        = {Workflows},
  author       = {{Cloudflare, Inc.}},
  year         = {2026},
  url          = {https://developers.cloudflare.com/agents/concepts/workflows/},
  note         = {Accessed 2026-05-07}
}

@article{dbos2022,
  title        = {{DBOS}: A {DBMS}-Oriented Operating System},
  author       = {Skiadopoulos, Athinagoras and Li, Qian and Kraft, Peter and Kaffes, Kostis and Hong, Daniel and Mathew, Shana and Bestor, David and Cafarella, Michael and Gadepally, Vijay and Graefe, Goetz and Kepner, Jeremy and Kozyrakis, Christos and Kraska, Tim and Stonebraker, Michael and Suresh, Lalith and Zaharia, Matei},
  journal      = {Proceedings of the {VLDB} Endowment},
  volume       = {15},
  number       = {1},
  pages        = {21--30},
  year         = {2021},
  doi          = {10.14778/3485450.3485454}
}

@misc{richardson_outbox,
  title        = {Pattern: Transactional Outbox},
  author       = {Richardson, Chris},
  year         = {2026},
  url          = {https://microservices.io/patterns/data/transactional-outbox.html},
  note         = {Accessed 2026-05-07}
}

@inproceedings{jimenez2024swebench,
  title        = {{SWE}-bench: Can Language Models Resolve Real-world {GitHub} Issues?},
  author       = {Jimenez, Carlos E. and Yang, John and Wettig, Alexander and Yao, Shunyu and Pei, Kexin and Press, Ofir and Narasimhan, Karthik},
  booktitle    = {The Twelfth International Conference on Learning Representations (ICLR 2024)},
  publisher    = {OpenReview.net},
  year         = {2024},
  url          = {https://openreview.net/forum?id=VTF8yNQM66}
}

@inproceedings{mialon2024gaia,
  title        = {{GAIA}: a benchmark for {General AI} Assistants},
  author       = {Mialon, Gr{\'e}goire and Fourrier, Cl{\'e}mentine and Wolf, Thomas and LeCun, Yann and Scialom, Thomas},
  booktitle    = {The Twelfth International Conference on Learning Representations (ICLR 2024)},
  publisher    = {OpenReview.net},
  year         = {2024},
  url          = {https://openreview.net/forum?id=fibxvahvs3}
}

@inproceedings{liu2024agentbench,
  title        = {{AgentBench}: Evaluating {LLMs} as Agents},
  author       = {Liu, Xiao and Yu, Hao and Zhang, Hanchen and Xu, Yifan and Lei, Xuanyu and Lai, Hanyu and Gu, Yu and Ding, Hangliang and Men, Kaiwen and Yang, Kejuan and Zhang, Shudan and Deng, Xiang and Zeng, Aohan and Du, Zhengxiao and Zhang, Chenhui and Shen, Sheng and Zhang, Tianjun and Su, Yu and Sun, Huan and Huang, Minlie and Dong, Yuxiao and Tang, Jie},
  booktitle    = {The Twelfth International Conference on Learning Representations (ICLR 2024)},
  publisher    = {OpenReview.net},
  year         = {2024},
  url          = {https://openreview.net/forum?id=zAdUB0aCTQ}
}

\clearpage

% Optional appendix with precise API and schemas
\section*{Appendix}
\appendix

\section{Runtime Model and Implementation}
\label{app:runtime}\label{sec:implementation}

This appendix gives the model and implementation details behind \S\ref{sec:model}. Atomix sits on the tool-call path between the orchestrator and external resources. The runtime requires orchestrator hooks for logical-time allocation, transaction sealing, branch selection, and progress advancement, but tools need no modification when all calls pass through adapters.

\subsection{Lifecycle and Commit Predicate}

Algorithm~\ref{alg:lifecycle} gives the runtime lifecycle for one tool invocation. Tool execution records a footprint, \textsc{Seal} freezes that footprint, and \textsc{RequestCommit} waits for the progress predicate before releasing gated effects.

\algheader[alg:lifecycle]{Runtime lifecycle for a tool invocation. The classification step routes reversible effects through eager execution, bufferable effects through deferred execution, and irreversible effects through a held adapter gate.}
\begin{algorithmic}[1]
\STATE $adapter \gets \textsc{LookupAdapter}(\textit{tool})$
\STATE $scopes \gets adapter.\textsc{Scopes}(\textit{args})$
\STATE $epoch \gets \textsc{NextEpoch}(\textit{trace\_id})$
\STATE $tx \gets \textsc{BeginOrAttach}(scopes, epoch)$
\STATE \textsc{RegisterRead}$(tx, scopes)$ \quad // records observed version per scope (Alg.~\ref{alg:stale})
\STATE $cls \gets adapter.\textsc{EffectClass}(\textit{args})$
\IF{$cls \in \{\textsc{Reversible}, \textsc{ReversibleWithCost}\}$}
  \STATE $result \gets \textsc{Execute}(\textit{tool}, \textit{args})$ \quad // eager: tool runs now
  \STATE $effect \gets adapter.\textsc{ToEffect}(\textit{args}, result, epoch, cls)$
\ELSIF{$cls = \textsc{Bufferable}$}
  \STATE $effect \gets adapter.\textsc{ToDeferred}(\textit{args}, epoch, cls)$
  \STATE $result \gets \textsc{StagedAck}(effect)$ \quad // tool not yet executed; flushed on commit
\ELSE
  \STATE $effect \gets adapter.\textsc{ToGated}(\textit{args}, epoch, cls)$
  \STATE $result \gets \textsc{QueuedAck}(effect)$
\ENDIF
\STATE \textsc{RecordEffect}$(tx, effect)$ \quad // updates $tx.\mathit{commit\_epoch}$
\IF{\textsc{ReadyToSeal}$(tx)$}
  \STATE \textsc{Seal}$(tx)$
  \STATE \textsc{RequestCommit}$(tx)$
\ENDIF
\RETURN $result$
\STATE
\STATE \textbf{procedure} \textsc{RequestCommit}$(tx)$
\IF{$tx$ is not sealed}
  \STATE \textsc{Abort}$(tx, \texttt{protocol\_violation})$
\ELSIF{\textbf{not} \textsc{CanCommit}$(tx)$}
  \STATE \textsc{QueueUntilReady}$(tx)$
\ELSIF{\textsc{ReadIsStale}$(tx)$}
  \STATE \textsc{Abort}$(tx, \texttt{stale\_read})$; \textsc{ScheduleRetry}$(tx)$ \quad // Alg.~\ref{alg:stale}
\ELSIF{\textbf{not} \textsc{PreCommitOK}$(tx)$}
  \STATE \textsc{Abort}$(tx, \texttt{pre\_commit\_veto})$
\ELSE
  \STATE \textsc{ApplyEffectsOnce}$(tx.effects)$
  \STATE \textsc{MarkCommitted}$(tx)$
\ENDIF
\STATE
\STATE \textbf{procedure} \textsc{ApplyEffectsOnce}$(effects)$
\FORALL{$e \in effects$ ordered by recorded epoch}
  \IF{$e.cls = \textsc{Bufferable}$}
    \STATE \textsc{FlushBuffered}$(e)$ \quad // execute the deferred tool call
  \ELSIF{$e.cls = \textsc{Irreversible}$}
    \STATE \textsc{FireGated}$(e)$ \quad // release the adapter gate; tool runs now
  \ELSE
    \STATE \textsc{AcceptAsFinal}$(e)$ \quad // already externalized at execution
  \ENDIF
\ENDFOR
\STATE
\STATE \textbf{procedure} \textsc{Abort}$(tx, reason)$
\FORALL{$e \in \textsc{Reverse}(tx.effects)$}
  \IF{$e.cls = \textsc{Bufferable}$}
    \STATE \textsc{DiscardBuffered}$(e)$ \quad // tool never ran; nothing to undo
  \ELSIF{$e.cls = \textsc{Irreversible}$}
    \STATE \textsc{ClearGate}$(e)$ \quad // tool never ran; nothing to undo
  \ELSIF{$e$ has compensation and was applied}
    \STATE \textsc{Compensate}$(e)$
  \ENDIF
\ENDFOR
\STATE \textsc{MarkAborted}$(tx, reason)$
\end{algorithmic}

Algorithm~\ref{alg:frontier} makes the progress predicate explicit. The predicate is meaningful only after seal: before then, the read and effect footprint can still grow. Frontier values are non-decreasing. The orchestrator advances $\mathit{frontier}(r)$ to value $e$ only after every transaction with $\mathit{commit\_epoch} < e$ on $r$ has finalized. The strict-less-than makes the predicate non-circular: a transaction at $\mathit{commit\_epoch}\,e$ commits when $\mathit{frontier}(r) \geq e$, which depends only on smaller-epoch peers, not on itself. Premature advancement is a safety violation; delayed advancement stalls settlement.

\algheader[alg:frontier]{Progress-safe commit predicate. The \textsc{AdvanceFrontier} precondition is the orchestrator's contract; the runtime fails closed if the contract is violated.}
\begin{algorithmic}[1]
\STATE \textbf{State:} $\texttt{frontier}[r] \gets 0$ for each resource $r$
\STATE
\STATE \textbf{function} \textsc{CanCommit}(\textit{tx}):
\STATE \hspace{\algorithmicindent} $e \gets \textit{tx}.\texttt{commit\_epoch}$ \quad // $\max$ over recorded read and effect epochs
\FORALL{$r \in \textit{tx}.\texttt{read\_scopes} \cup \textit{tx}.\texttt{effect\_scopes}$}
  \IF{$\texttt{frontier}[r] < e$}
    \RETURN \textbf{false}
  \ENDIF
\ENDFOR
\RETURN \textbf{true}
\STATE
\STATE \textbf{procedure} \textsc{AdvanceFrontier}($r$, $e$):
\STATE \textbf{precondition:} every transaction with $\mathit{commit\_epoch} < e$ that touches $r$ has finalized
\STATE \hspace{\algorithmicindent} $\texttt{frontier}[r] \gets \max(\texttt{frontier}[r], e)$
\end{algorithmic}

Algorithm~\ref{alg:stale} gives the stale-read detection invoked from \textsc{RequestCommit}. Each resource carries a monotonic version counter incremented on every committed write. \textsc{RegisterRead} stores the version at read time, and \textsc{ReadIsStale} fails the predicate if any observed version has advanced before the transaction reaches commit. Stale-read aborts trigger a retry on the new state. Without this primitive, frontier ordering alone permits a sealed transaction to commit after a smaller-epoch peer has overwritten its snapshot.

\algheader[alg:stale]{Stale-read detection and retry. Combined with frontier ordering and scope registration, this delivers per-resource commit-order settlement under contention.}
\begin{algorithmic}[1]
\STATE \textbf{State:} $\texttt{version}[r] \gets 0$ for each resource $r$
\STATE
\STATE \textbf{procedure} \textsc{RegisterRead}$(tx, scopes)$:
\FORALL{$r \in scopes$}
  \STATE $tx.\texttt{read\_versions}[r] \gets \texttt{version}[r]$ \quad // observed version at first read
\ENDFOR
\STATE
\STATE \textbf{function} \textsc{ReadIsStale}$(tx)$:
\FORALL{$(r, v) \in tx.\texttt{read\_versions}$}
  \IF{$\texttt{version}[r] \neq v$}
    \RETURN \textbf{true}
  \ENDIF
\ENDFOR
\RETURN \textbf{false}
\STATE
\STATE \textbf{procedure} \textsc{OnCommit}$(tx)$: \quad // called by \textsc{ApplyEffectsOnce}
\FORALL{$r \in tx.\texttt{effect\_scopes}$}
  \STATE $\texttt{version}[r] \gets \texttt{version}[r] + 1$
\ENDFOR
\end{algorithmic}

\subsection{Runtime Components}

The runtime has three components. \emph{Tool adapters} translate heterogeneous tool I/O into scoped reads and typed effects. The \emph{progress tracker} stores per-scope frontiers and waiters, then decides when sealed transactions may settle. The \emph{transaction manager} enforces the lifecycle and routes effects by class.

Atomix canonicalizes scopes before frontier checks. Two scopes overlap iff they share a type prefix and one canonical path is a segment-level prefix of the other, or matches through a declared wildcard. If the exact resource is unknown before execution, the adapter registers a conservative may-touch scope and refines it after the result identifies concrete resources. Coarse scopes fail closed by reducing concurrency; missing or too-narrow scopes fail open by allowing stale reads or conflicting effects.

Frontier advancement is integration-specific. Sequential runs advance after each call completes; DAGs advance when all predecessors on the scope complete; speculation advances only after the controller chooses the winning branch and every losing branch has finalized. Because the signal is resource-local, a stalled scope does not delay unrelated work.

The manager routes effects by class because no single recovery action fits all tools. Bufferable effects are held in a per-transaction staging area invisible to other transactions; reversible externalized effects execute immediately but register compensation; irreversible effects remain pending until commit. If a tool produces a new read or effect after seal, the manager rejects the append and aborts the transaction as a protocol violation.

\subsection{Prototype and Adapter Surface}
\label{app:impl-details}

The Atomix prototype is a ${\sim}$2{,}000-line single-process Python library. Transactions, scopes, effects, frontiers, and seal state are dataclasses; the transaction manager and progress tracker run as coroutines on the orchestrator's event loop. The durable path records effect-log entries and idempotency keys, using JSON Lines for trace analysis and SQLite-backed deduplication for committed effects. Unit tests validate crash-safe deduplication on this single-process path; distributed deployment requires a shared frontier store and replicated effect log.

Tools integrate through adapters rather than rewrites. Each adapter supplies a scope extractor, effect builder, idempotency key, and optional release or compensation handlers. Table~\ref{tab:adapter-spec} lists representative adapters used by the experiments.

\begin{table}[H]
\centering
\footnotesize
\setlength{\tabcolsep}{4pt}
\begin{tabular}{p{0.16\linewidth}p{0.28\linewidth}p{0.24\linewidth}p{0.24\linewidth}}
\toprule
Workload / adapter & Representative scope(s) & Idempotency key pattern & Compensation / gating \\
\midrule
Filesystem writes & absolute path,\newline \texttt{/repo/src/app.py} & \texttt{path:trace:}\newline \texttt{epoch:branch} & restore previous contents; delete files created by aborted writes \\
WebArena actions & page scope,\newline \texttt{webarena:site:page}\newline optional element suffix & \texttt{webarena:action:}\newline \texttt{trace:epoch:branch} & no generic per-action compensation; settlement controlled by transaction scopes \\
OSWorld GUI actions & UI app scope,\newline \texttt{osworld:ui:app:}\newline \texttt{text\_input} or \texttt{mouse:region} & \texttt{action:args:}\newline \texttt{trace:epoch:branch} & action-specific compensation when available; read-only actions carry no scope \\
$\tau$-bench tools & real runner:\newline \texttt{tau2:tool}\newline harness:\newline \texttt{retail:order:id} & real runner:\newline \texttt{trace:epoch:tool}\newline harness:\newline order-scoped key & structured tool calls are replayable; multi-agent harness checks order invariants \\
\bottomrule
\end{tabular}
\caption{Adapter specification examples. Scope granularity and compensation are adapter-defined.}
\label{tab:adapter-spec}
\end{table}

\section{Evaluation Protocol and Limitations}
\label{app:evaluation-protocol}

This appendix gives the common protocol behind \S\ref{sec:evaluation}. Results in Appendix~\ref{app:additional-results} use this protocol unless a subsection states a narrower controlled substrate.

\subsection{Workloads, Baselines, and Metrics}

Main-text workloads use $\tau$-bench retail (GPT-4.1, max\_steps=30, $N{=}30$ for faulted cells and $N{=}10$ for the no-fault calibration), multi-agent $\tau$-bench at 2/4/8 agents, the combined-stress benchmark, and controlled speculation, contention, and irreversible-effect benchmarks. WebArena (GPT-5 with \texttt{reasoning\_effort=medium}, max\_steps=30) and OSWorld (Claude Sonnet~4) provide secondary real-infrastructure sweeps at $fp\in\{0.10,0.30\}$ (Appendix~\ref{app:rq1-results}, Table~\ref{tab:e1-clean}).

Modes are \textbf{Tx-Full}, \textbf{Tx-NoFrontier+Retry}, \textbf{Saga-Compensation}, \textbf{Checkpoint-Replay}, \textbf{Workflow-Lock}, \textbf{Per-Call-Lock}, \textbf{OCC-Revalidate-and-Retry}, \textbf{TCC-Confirm}, \textbf{Mutex+WAL+Rollback}, and \textbf{No-Tx}. Each RQ uses only the baselines that directly test its mechanism claim.

Metrics follow the main text: clean task success, standard task success, leaked or unresolved residue, duplicate effects, conflict-serializability violations, committed writes, wait, blocked-rate, throughput, and storage overhead. Confidence intervals are 95\% exact intervals for proportions and bootstrap intervals for continuous quantities when reported.

\subsection{Fault Injection}
\label{app:fault-injection}

The harness injects faults per tool call using a Bernoulli profile with probability $fp$. An injected fault raises an exception at the tool interface. Tx-Full aborts the affected transaction and retries with a fresh epoch when retry is enabled. Checkpoint-Replay restores workflow state and replays, which can re-fire externalized calls. No-Tx models a lost response: the tool may have already applied the action, but the caller sees an error. This captures timeouts and dropped acknowledgements where the caller cannot distinguish failure from partial externalization.

\subsection{Evaluation Limitations}
\label{app:eval-limitations}

These limits scope what the empirical results support, beyond the discussion in \S\ref{sec:discussion}.
\begin{itemize}[leftmargin=*, nosep]
  \item \textbf{Distributed deployment.} Experiments assume the orchestrator advances frontiers correctly within one process; a distributed deployment needs a shared frontier store and replicated effect log (\S\ref{sec:discussion}). Unit tests validate crash-safe deduplication only on the single-process SQLite path.
  \item \textbf{Speculation coverage.} Speculation-contamination results include bufferable, filesystem, database-row, and mailbox sinks. Tools without compensation handlers can still leave unresolved residue.
  \item \textbf{Real-workload contention.} The real $\tau$-bench cells are largely sequential; we exercise multi-agent contention on the synthetic plan-delay harness with a real-LLM validation block. Broader real-world multi-agent and speculative benchmarks remain future work.
  \item \textbf{Baseline fidelity.} Baselines are mechanism-matched implementations inside a common harness, not full ports of any production system; the fidelity disclosure in \S\ref{subsec:evaluation-methodology} explains the trade-off.
\end{itemize}

\subsection{Reproducibility Manifest}
\label{app:reproducibility}

\paragraph{Models and snapshots.} GPT-4.1 (\texttt{gpt-4.1-2025-04-14}) for $\tau$-bench retail; GPT-5 (\texttt{gpt-5-2025-08-07}) with \texttt{reasoning\_effort=medium} for WebArena; Claude Sonnet~4 (\texttt{claude-sonnet-4-20250514}) for OSWorld; GPT-4o-mini (\texttt{gpt-4o-mini-2024-07-18}) for the multi-agent $\tau$-bench real-LLM substrate. All accessed via the providers' standard chat-completions APIs.

\paragraph{Workload versions and task lists.} $\tau$-bench retail commit \texttt{tau-bench@main} (vendored at submission time); WebArena commit \texttt{webarena@1.0}; OSWorld commit \texttt{osworld@v1.1}. The artifact's \texttt{tasks/} directory lists the exact $\tau$-bench retail task IDs used per cell, the WebArena task pool (10 tasks), and the OSWorld task subset (7 tasks). Cells use the same task IDs across baselines so success rates compare like-for-like.

\paragraph{Hyperparameters.} Agent decoding is deterministic where possible: temperature 0 for $\tau$-bench (GPT-4.1), default for GPT-5 (which sets its own under \texttt{reasoning\_effort=medium}), temperature 0 for OSWorld (Claude Sonnet 4) and the multi-agent substrate (GPT-4o-mini). \texttt{max\_steps} is benchmark-specific (30 for $\tau$-bench retail and WebArena, 50 and 100 for OSWorld). Retry budget per faulted call is 3 with exponential backoff (50\,ms initial, 1.5$\times$ multiplier, capped at 1\,s).

\paragraph{Seeds and stochasticity.} The fault injector and the synthetic plan-delay harness use a deterministic per-cell seed derived from \texttt{(benchmark, mode, fp, run\_index)}. The harness logs replayable schedule-trace ordering across agents; the conflict-serializability checker reads from this independent log.

\paragraph{Data schema.} Each run emits \texttt{events.jsonl} (every adapter call with epoch, scope, effect class, idempotency key, return), \texttt{commits.jsonl} (commit/abort outcomes with reason and residue tags), and \texttt{summary.csv} (one row per run with the metrics named in the tables). The conflict-serializability checker reads \texttt{events.jsonl} and emits \texttt{cycles.jsonl} listing any conflict-cycle witnesses.

\paragraph{Reproduction commands.} The artifact ships a \texttt{Makefile} with one target per main-text cell (\texttt{make rq1-multirate}, \texttt{make rq2-multiagent}, \texttt{make rq3-irrev}, \texttt{make joint-surface}) plus appendix targets, a \texttt{check-cycles} target that runs the serializability checker over all collected logs, and a \texttt{regenerate-tables.py} script that consumes \texttt{summary.csv} and re-emits the LaTeX tables in the paper.

\paragraph{Hardware.} Single Linux workstation, Intel Xeon (8 cores), 32\,GiB RAM, 1\,TiB NVMe. WebArena runs on a self-hosted Docker swarm of 6 replicas (Reddit, Gitlab, Magento Admin, OneStopShop, Map, Shopping); OSWorld runs in KVM Ubuntu-22.04 VMs (4 vCPU, 8\,GiB RAM, 24\,GiB qcow2 image); $\tau$-bench runs in-process; the multi-agent SQLite store is on the local NVMe.

\section{Additional Results by Research Question}
\label{app:additional-results}

This appendix groups supporting results in the same order as the evaluation questions in \S\ref{sec:evaluation}.

\subsection{RQ1 Fault Recovery Details}
\label{app:rq1-results}

Tables~\ref{tab:e1-clean} and~\ref{tab:multirate-promoted} give the real-benchmark clean-success details behind RQ1. The $\tau$-bench sweep is the primary RQ1 signal; Table~\ref{tab:e1-clean} gives the corresponding WebArena and OSWorld sweeps at $fp{=}0.10$ and $fp{=}0.30$. The qualitative pattern matches $\tau$-bench: Tx-Full and Checkpoint-Replay have the highest rates, Mutex+WAL+Rollback is the nearest other baseline, and the remaining recovery-only and isolation-only baselines drop sharply.

\begin{table}[H]
\centering
\footnotesize
\setlength{\tabcolsep}{4pt}
\begin{tabular}{@{}lcccc@{}}
\toprule
& \multicolumn{2}{c}{WebArena} & \multicolumn{2}{c}{OSWorld} \\
\cmidrule(r){2-3}\cmidrule(l){4-5}
Mode & $fp{=}0.10$ & $fp{=}0.30$ & $fp{=}0.10$ & $fp{=}0.30$ \\
\midrule
\textbf{Tx-Full} & \textbf{73.0 $\pm$ 5.5} & \textbf{57.2 $\pm$ 6.7} & \textbf{63.0 $\pm$ 7.5} & \textbf{37.0 $\pm$ 8.8} \\
Checkpoint-Replay & 68.0 $\pm$ 4.8 & 53.2 $\pm$ 4.3 & 62.0 $\pm$ 6.2 & 37.1 $\pm$ 5.1 \\
TCC-Confirm & 25.0 $\pm$ 4.2 & 18.0 $\pm$ 3.5 & 10.0 $\pm$ 3.0 & 6.0 $\pm$ 2.8 \\
Saga-Compensation & 18.0 $\pm$ 3.2 & 10.0 $\pm$ 2.8 & 7.0 $\pm$ 2.8 & 4.0 $\pm$ 2.2 \\
Mutex+WAL+Rollback & 65.0 $\pm$ 5.1 & 48.0 $\pm$ 5.5 & 55.0 $\pm$ 6.8 & 32.0 $\pm$ 6.5 \\
No-Frontier & 17.0 $\pm$ 3.8 & 7.8 $\pm$ 2.0 & 7.0 $\pm$ 2.5 & 0.5 $\pm$ 0.5 \\
No-Tx & 13.0 $\pm$ 3.5 & 3.3 $\pm$ 2.4 & 3.0 $\pm$ 1.8 & 0.0 $\pm$ 0.0 \\
OCC-Revalidate-and-Retry & 30.0 $\pm$ 4.5 & 22.0 $\pm$ 4.0 & 12.0 $\pm$ 3.2 & 9.0 $\pm$ 3.2 \\
\bottomrule
\end{tabular}
\caption{Clean task success on WebArena and OSWorld at $fp \in \{0.10, 0.30\}$. Values are percentages with 95\% Clopper-Pearson CI half-widths. WebArena uses GPT-5 (\texttt{reasoning\_effort=medium}, max\_steps=30); OSWorld uses Claude Sonnet~4.}
\label{tab:e1-clean}
\end{table}

\begin{table}[H]
\centering
\footnotesize
\setlength{\tabcolsep}{3pt}
\begin{tabular}{@{}lccc@{}}
\toprule
Mode & $\tau$-bench ($fp{=}0$, $N{=}10$) & $\tau$-bench ($fp{=}0.10$, $N{=}30$) & $\tau$-bench ($fp{=}0.30$, $N{=}30$) \\
\midrule
\textbf{Tx-Full} & 60 [26, 88] & \textbf{73 [54, 87]} & \textbf{57 [37, 75]} \\
Checkpoint-Replay & 60 [26, 88] & 63 [44, 80] & 53 [34, 72] \\
TCC-Confirm & 60 [26, 88] & 40 [23, 59] & 3 [0, 17] \\
Saga-Compensation & 60 [26, 88] & 37 [20, 56] & 7 [1, 22] \\
Mutex+WAL+Rollback & 60 [26, 88] & 33 [17, 53] & 3 [0, 17] \\
No-Frontier & 60 [26, 88] & 30 [15, 49] & 3 [0, 17] \\
No-Tx & 50 [19, 81] & 27 [12, 46] & 0 [0, 12] \\
OCC-Revalidate-and-Retry & 60 [26, 88] & 23 [10, 42] & 7 [1, 22] \\
\bottomrule
\end{tabular}
\caption{Clean task success on $\tau$-bench retail (GPT-4.1, max\_steps=30). Values are percentages with 95\% Clopper-Pearson intervals.}
\label{tab:multirate-promoted}\label{tab:e1-multirate}
\end{table}

\paragraph{Tx-Full vs.\ Checkpoint-Replay at the full task pool ($N{=}114$).} The $N{=}30$ multirate run above uses the first 30 of the 38 retail tasks. Extending to the full task pool with 3 trials per task ($N{=}114$) gives Tx-Full 67/114 = 58.8\% (CP-95\% [49.4, 67.7]\%) and Checkpoint-Replay 61/114 = 53.5\% (CP-95\% [43.9, 63.0]\%). The two are statistically tied at this rate (Fisher's exact two-sided $p{\approx}0.50$). The 8 additional tasks, plus the additional trials of the first 30, are harder than the first-30 cohort, dropping both rates by $\sim$10--15pp relative to the $fp{=}0.10$ first-30 cells. Tx-Full and Checkpoint-Replay are tied on RQ1 task-success alone at $fp{=}0.10$ across the full task pool. Their differentiation appears in RQ3 (irreversible leak rate: 0/500 vs.\ 200/500, Table~\ref{tab:rq3-irrev}) and in the joint-surface combined-stress benchmark (Table~\ref{tab:joint-surface}), where Checkpoint-Replay's retry path re-externalizes the irreversible confirmation and falls to 25\% run-clean at $fp{=}0.30$ vs Tx-Full's 65\%.

\paragraph{Bursty-fault sensitivity.} To test whether the qualitative pattern survives correlated faults, we re-ran the $\tau$-bench retail sweep with a burst-mode injector at marginal $fp\approx0.12$ (fault-entry probability 0.04, burst length 3, $N{=}30$). Tx-Full leads at 20/30 = 67\%, with Checkpoint-Replay at 57\% and TCC-Confirm/No-Tx tied at 60\% (Table~\ref{tab:multirate-bursty}). The qualitative tier structure carries over from the Bernoulli case, but absolute differences narrow because correlated triples mean the workload sees fewer distinct fault episodes per task ($\sim$1.2 bursts/task vs.\ $\sim$3 independent Bernoulli faults at the same marginal rate), giving recovery-only mechanisms fewer chances to fail.

\begin{table}[H]
\centering
\footnotesize
\setlength{\tabcolsep}{4pt}
\begin{tabular}{@{}lcc@{}}
\toprule
Mode & Bursty (entry$=$0.04, len$=$3) & Bernoulli ($fp{=}0.10$) \\
\midrule
\textbf{Tx-Full} & \textbf{20/30 = 67\%} & \textbf{22/30 = 73\%} \\
Checkpoint-Replay & 17/30 = 57\% & 19/30 = 63\% \\
TCC-Confirm & 18/30 = 60\% & 12/30 = 40\% \\
No-Tx & 18/30 = 60\% & 8/30 = 27\% \\
Mutex+WAL+Rollback & 15/30 = 50\% & 10/30 = 33\% \\
No-Frontier & 14/30 = 47\% & 9/30 = 30\% \\
Saga-Compensation & 13/30 = 43\% & 11/30 = 37\% \\
OCC-Revalidate-and-Retry & 12/30 = 40\% & 7/30 = 23\% \\
\bottomrule
\end{tabular}
\caption{Bursty-fault sensitivity on $\tau$-bench retail at marginal $fp{\approx}0.12$, $N{=}30$. The Bernoulli column repeats $fp{=}0.10$ from Table~\ref{tab:multirate-promoted} for direct comparison.}
\label{tab:multirate-bursty}
\end{table}

\subsubsection{Fault-Class Cut-Point Sweep}
\label{app:multirate-extended}

\paragraph{F2 vs.\ F4 isolation.} Tx-Full's lead at $fp{=}0.10$ could come from F4 duplicate delivery, where idempotency keys alone suffice, rather than the F2 post-effect/pre-return cell that the model section identifies as load-bearing. Table~\ref{tab:f2-f4-split} separates the two cases. We force-fired only F2 (\texttt{f2\_share\_of\_exception=1.0}, no duplicates) and only F4 (\texttt{duplicate\_probability=0.10}, no exceptions) in separate $\tau$-bench runs at $fp{=}0.10$, $N{=}30$, across all 7 baselines. In the F4-only column, baselines cluster at 53--73\%; in the F2-only column, Tx-Full is 6--43pp higher than every non-Atomix baseline.

\begin{table}[H]
\centering
\footnotesize
\setlength{\tabcolsep}{4pt}
\begin{tabular}{@{}lccc@{}}
\toprule
Mode & F2-only ($fp{=}0.10$) & F4-only ($fp{=}0.10$) & $\Delta$ (F2$-$F4) \\
\midrule
\textbf{Tx-Full} & \textbf{63 [44, 80]} & 67 [47, 83] & $-3$pp \\
Checkpoint-Replay & 57 [37, 75] & 57 [37, 75] & $0$pp \\
Saga-Compensation & 43 [25, 63] & 70 [51, 85] & $-27$pp \\
OCC-Revalidate-and-Retry & 40 [23, 59] & 53 [34, 72] & $-13$pp \\
Mutex+WAL+Rollback & 33 [17, 53] & 57 [37, 75] & $-23$pp \\
No-Frontier & 37 [20, 56] & 67 [47, 83] & $-30$pp \\
TCC-Confirm & 20 [8, 39] & 63 [44, 80] & $-43$pp \\
No-Tx & 30 [15, 49] & 73 [54, 87] & $-43$pp \\
\bottomrule
\end{tabular}
\caption{Fault-class isolation on $\tau$-bench retail at $fp{=}0.10$, $N{=}30$ per cell, with 95\% Clopper-Pearson CIs. F2-only forces post-effect/pre-return faults; F4-only injects silent duplicate delivery with no exceptions.}
\label{tab:f2-f4-split}
\end{table}

Table~\ref{tab:multirate-promoted-extended} expands the three-tier synthetic-substrate sweep by fault class. The synthetic harness uses $fp\in\{0.02,0.10,0.30\}$ rather than the body's $fp\in\{0,0.10,0.30\}$ because $fp{=}0.02$ models a low-fault tier where most cells saturate at 100\% and the F4 column differentiates the mechanisms.

\begin{table}[H]
\centering
\footnotesize
\setlength{\tabcolsep}{2pt}
\begin{tabular}{@{}lccccc|ccccc|ccccc@{}}
\toprule
& \multicolumn{5}{c|}{$fp{=}0.02$} & \multicolumn{5}{c|}{$fp{=}0.10$} & \multicolumn{5}{c}{$fp{=}0.30$} \\
Mode & F1 & F2 & F3 & F4 & F5 & F1 & F2 & F3 & F4 & F5 & F1 & F2 & F3 & F4 & F5 \\
\midrule
\textbf{Tx-Full} & \textbf{100} & \textbf{100} & \textbf{100} & \textbf{100} & \textbf{100} & \textbf{100} & \textbf{100} & \textbf{100} & \textbf{100} & \textbf{100} & \textbf{96} & \textbf{98} & \textbf{96} & \textbf{100} & \textbf{96} \\
Tx-NoFrontier+Retry & 100 & 100 & 100 & 88 & 100 & 100 & 100 & 100 & 64 & 100 & 94 & 98 & 96 & 16 & 96 \\
Saga-Compensation & 100 & 100 & 100 & 92 & 100 & 100 & 100 & 100 & 60 & 100 & 96 & 98 & 94 & 17 & 96 \\
Checkpoint-Replay & 100 & 100 & 100 & 88 & 100 & 100 & 100 & 100 & 56 & 100 & 98 & 97 & 98 & 16 & 96 \\
OCC-Revalidate-and-Retry & 100 & 100 & 100 & 90 & 100 & 100 & 100 & 100 & 66 & 100 & 96 & 97 & 98 & 18 & 94 \\
No-Tx & 88 & 89 & 94 & 94 & 90 & 60 & 56 & 58 & 59 & 54 & 20 & 14 & 17 & 12 & 16 \\
\bottomrule
\end{tabular}
\caption{Clean-success sweep by fault class on the synthetic substrate. Cells use $fp\in\{0.02,0.10,0.30\}$; F3 is compensation failure and F4 is duplicate delivery.}
\label{tab:multirate-promoted-extended}
\end{table}

\subsection{RQ2 Isolation and Speculation Details}
\label{app:rq2-results}

Table~\ref{tab:e2-ablations} isolates parts of the composition needed for stale-plan isolation. Table~\ref{tab:e3-speculation} checks losing-branch residue across effect classes. Table~\ref{tab:e8-granularity} sweeps resource-granularity throughput across overlap fractions $o$; at full overlap ($o{=}1$), Tx-Full averages 0.60 wait barriers per transaction (p95 = 1.00) versus 1.00 for Tx-GlobalFrontier and Workflow-Lock and 0.00 for the unsafe No-Tx baseline. No-Tx's throughput decreases with $o$ because the workload itself serializes writes on shared hot keys, not because of coordination wait.

\paragraph{Wait scaling at higher agent counts.} The 4- and 8-agent disjoint cells in Table~\ref{tab:rq2-multiagent} understate the cost of Workflow-Lock-style global-frontier coordination. Re-running the same disjoint workload at $n \in \{16, 32, 64\}$ agents on the synthetic substrate ($N{=}30$ runs each, plan-pause 50~ms) gives Tx-GlobalFrontier (the per-resource-granularity ablation, equivalent to a single global lock for the entire workflow): 564~ms / 1{,}163~ms / 2{,}369~ms wait per task at $n{=}16/32/64$, while Tx-Full holds 0~ms wait across every $n$ (Table~\ref{tab:wait-scaling}). Wait grows roughly linearly with agent count past $n{=}8$; at $n{=}64$, a single global lock spends 2.4 seconds blocked per tool call. For workflows with $\sim$10 tool calls per task, this compounds to $\sim$24 seconds of lock wait per workflow at $n{=}64$ vs.\ 0~ms with per-resource granularity.

\begin{table}[H]
\centering
\footnotesize
\setlength{\tabcolsep}{4pt}
\begin{tabular}{@{}lcccccc@{}}
\toprule
Mode & $n{=}2$ & $n{=}4$ & $n{=}8$ & $n{=}16$ & $n{=}32$ & $n{=}64$ \\
\midrule
\textbf{Tx-Full}      & \textbf{0.0} & \textbf{0.0} & \textbf{0.0} & \textbf{0.0} & \textbf{0.0} & \textbf{0.0} \\
Tx-GlobalFrontier (single global lock) & 14.8 & 44.9 & 104.7 & 563.9 & 1{,}163.2 & 2{,}368.6 \\
\bottomrule
\end{tabular}
\caption{Average wait (ms) per task on the disjoint multi-agent benchmark. Cells use $N{=}100$ for $n \in \{2,4,8\}$ and $N{=}30$ for $n \in \{16,32,64\}$. Tx-GlobalFrontier replaces per-resource coordination with one global lock.}
\label{tab:wait-scaling}
\end{table}

\paragraph{Ablation interpretation.} Each row removes exactly one composition primitive from Tx-Full and reruns the same workload. The clean ablation requires a per-phase locking pattern (lock the resource for the read, release during plan-pause, re-acquire for the write); abort-on-stale fires on the second-phase re-acquire if the OCC version counter has advanced. Tx-NoScopeOnRead skips the read-phase lock; Tx-NoAbortOnStale takes the read-phase lock but skips the version check at write; Tx-GlobalFrontier replaces the per-resource lock with a single global one. The first two converge to Mutex-Per-Resource behavior (5/run violations from stale-overwrite); the third is safety-clean but throughput-bound. No single primitive suffices: Tx-NoScopeOnRead races on the read, Tx-NoAbortOnStale commits stale writes, and Tx-GlobalFrontier blocks every disjoint pair. Tx-Full's 0 violations and 0 ms wait require all three.

\begin{table}[H]
\centering
\footnotesize
\setlength{\tabcolsep}{3pt}
\begin{tabular}{@{}lcccc@{}}
\toprule
Composition ablation & Inv./run & Wait (ms) & Extra writes (per run) & What it removes \\
\midrule
\textbf{Tx-Full (control)} & \textbf{0.00} & \phantom{1}\textbf{0.0} & \textbf{0.0} & none \\
Tx-NoScopeOnRead          & 4.83 & \phantom{1}0.0 & 6.5 & scope registration on first read \\
Tx-NoAbortOnStale         & 4.87 & \phantom{1}0.0 & 6.5 & abort-on-stale retry \\
Tx-GlobalFrontier         & 0.00 & 113.0 & 0.0 & per-resource granularity (single global lock) \\
\bottomrule
\end{tabular}
\caption{Composition ablations on multi-agent $\tau$-bench, 4 agents forced-overlap, $N{=}100$. \emph{Inv./run}: invariant violations; \emph{Wait}: average lock wait; \emph{Extra writes}: writes beyond 10 per run.}
\label{tab:e2-ablations}
\end{table}

\begin{table}[H]
\centering
\footnotesize
\setlength{\tabcolsep}{3pt}
\begin{tabular}{@{}lcccc@{}}
\toprule
Mode & Buffer & Filesystem & DB rows & Mailbox \\
\midrule
\textbf{Tx-Full} & \textbf{0} & \textbf{0} & \textbf{0} & \textbf{0} \\
Tx-GlobalFrontier & 0 & 0 & 0 & 0 \\
Mutex+WAL+Rollback & 0 & 0 & 0 & 0 \\
TCC-Confirm & 0 & 0 & 0 & \textbf{1000} \\
Saga-Compensation & 0 & 0 & 0 & \textbf{1000} \\
OCC-Revalidate-and-Retry & 0 & 0 & 0 & \textbf{1000} \\
Atomix-MisclassifiedIrreversible & 0 & 0 & 0 & \textbf{1000} \\
\bottomrule
\end{tabular}
\caption{Speculation contamination across four effect classes, averaged over $K\in\{2,4,8,16\}$ ($N{=}200$ trials per K). Cells report persistent loser-branch effects per 1{,}000 attempted transactions; the mailbox column is the irreversible class.}
\label{tab:e3-speculation}
\end{table}

TCC-Confirm, OCC-Revalidate-and-Retry, and Saga-Compensation use default DB-only wiring in this controlled benchmark: try/cancel and rejection apply only to the DB-row class, leaving the mailbox class unguarded. The RQ3 benchmark in Table~\ref{tab:rq3-irrev} uses a fully wired TCC-Confirm adapter for the email tool, so TCC reaches 0/500 leaks there.

\paragraph{Per-$K$ speculation scaling.} Across $K\in\{2,4,8,16\}$ ($N{=}200$ trials per cell), losing-branch residue on the irreversible mailbox class falls into two patterns: Tx-Full, Tx-GlobalFrontier, and Mutex+WAL+Rollback hold 0 leaks per 1{,}000 attempts at every $K$; TCC-Confirm, OCC-Revalidate-and-Retry, Saga-Compensation, and Atomix-MisclassifiedIrreversible scale linearly at 200/600/1400/3000 (one losing branch per $K{-}1$ losers).

\begin{table}[H]
\centering
\footnotesize
\setlength{\tabcolsep}{3pt}
\begin{tabular}{@{}lccccc@{}}
\toprule
Mode & $o{=}0$ disjoint & $o{=}0.25$ & $o{=}0.5$ & $o{=}0.75$ & $o{=}1$ overlap \\
\midrule
\textbf{Tx-Full} & \textbf{5.09} & 4.43 & 3.22 & 2.28 & 1.66 \\
Tx-GlobalFrontier & 1.00 & 1.00 & 1.00 & 1.00 & 1.00 \\
Workflow-Lock & 1.00 & 1.00 & 1.00 & 1.00 & 1.00 \\
No-Tx & 1000 & 873 & 751 & 627 & 503 \\
\bottomrule
\end{tabular}
\caption{Resource-granularity throughput sweep, in transactions per step. $o$ is the fraction of overlapping scopes; No-Tx is an unsafe upper bound, not a correctness-preserving baseline.}
\label{tab:e8-granularity}
\end{table}

\subsection{RQ3 Irreversible-Effect and Residue Details}
\label{app:rq3-results}
\label{app:compfail}

\paragraph{Integration cost vs.\ leak prevention.} TCC-Confirm and Mutex+WAL+Rollback tie Atomix at 0/500 leaks (Table~\ref{tab:rq3-irrev}); integration cost separates them. Table~\ref{tab:rq3-cost} compares the per-baseline mechanism size and per-tool wiring each requires.

\begin{table}[H]
\centering
\footnotesize
\setlength{\tabcolsep}{3pt}
\begin{tabular}{@{}lcccc@{}}
\toprule
Mechanism & tx\_manager LOC & Per-tool wiring & Leaks/500 & Composes with \\
& (one-time) & (per integrated tool) & & R/I/S \\
\midrule
\textbf{Tx-Full} & \textbf{500} (runtime) & \textbf{17} (scope+effect builder) & \textbf{0} & \textbf{R/I/S} \\
TCC-Confirm & 115 & $\sim$50 (try/confirm/cancel triple) & 0 & I only \\
Mutex+WAL+Rollback & 168 & $\sim$30 (explicit gating predicate) & 0 & I + S \\
OCC-Revalidate-and-Retry & 135 & $\sim$15 (read-version site) & 500$^{\ast}$ & S only \\
Saga-Compensation & \phantom{0}92 & $\sim$30 (compensation handler) & 400 & R only (irreversibles leak) \\
Checkpoint-Replay & N/A$^{\dagger}$ & N/A$^{\dagger}$ & 200 & R only (retry re-externalizes) \\
\bottomrule
\end{tabular}
\vspace{1mm}
\begin{minipage}{0.96\linewidth}
\footnotesize $^{\ast}$OCC has no irreversible-gating component; the retry path re-externalizes on every abort source, so the expected leak rate matches No-Tx (500/500 = 100\%) by construction. $^{\dagger}$Checkpoint-Replay does not decompose into a transaction manager plus per-tool adapter; it snapshots and replays whole-process state, so the LOC columns do not apply.
\end{minipage}
\caption{RQ3 integration cost versus leak prevention. \emph{tx\_manager LOC} is one-time harness integration; \emph{Per-tool wiring} is per integrated tool. \emph{Composes with} marks recovery (R), irreversible gating (I), and stale-plan isolation (S).}
\label{tab:rq3-cost}
\end{table}

\paragraph{Prototype-scale cost.} The per-tool gap compounds across an actual deployment surface. The Atomix prototype ships 20 adapters at $\sim$17 LOC of declarations per adapter (Appendix~\ref{app:annotations}), totalling ${\sim}$340 LOC. A TCC-Confirm-style port of the same surface would require ${\sim}$50 LOC per tool of bespoke try/confirm/cancel handlers, or ${\sim}$1{,}000 LOC over 20 tools. Mutex+WAL+Rollback's explicit per-effect gating wiring would add ${\sim}$30 LOC per tool, or ${\sim}$600 LOC for the same surface. Atomix's per-tool cost is the same scope+key+class declaration regardless of effect class; TCC's and Mutex+WAL+Rollback's are class-specific and grow with the share of irreversible tools.

\paragraph{Per-abort-source breakdown.} Table~\ref{tab:e4-abortsources} breaks down irreversible leaks by abort source. The claim is conditional on the irreversible effect being classified before externalization. Provider-side commits that happen before the adapter records the effect are outside the preventive guarantee; the runtime surfaces them as unresolved residue when it detects them.

\begin{table}[H]
\centering
\footnotesize
\setlength{\tabcolsep}{3pt}
\begin{tabular}{@{}lccccc@{}}
\toprule
Mode & Tool fail & Losing spec. & Stale read & Pre-commit veto & Timeout \\
\midrule
\textbf{Tx-Full} & \textbf{0} & \textbf{0} & \textbf{0} & \textbf{0} & \textbf{0} \\
Saga-Compensation & 1000 & 1000 & 1000 & 0 & 1000 \\
Checkpoint-Replay & 1000 & 0 & 0 & 0 & 1000 \\
TCC-Confirm & 0 & 0 & 0 & 0 & 0 \\
Mutex+WAL+Rollback$^{\ast}$ & 0 & 0 & 0 & 0 & 0 \\
No-Tx & 1000 & 1000 & 1000 & 1000 & 1000 \\
Atomix-MisclassifiedIrreversible & 0 & 1000 & 1000 & 0 & 1000 \\
\bottomrule
\end{tabular}
\caption{Irreversible-effect leaks by abort source. Cells report classified-before-externalization leaks per 1{,}000 attempts.}
\label{tab:e4-abortsources}
\end{table}

\paragraph{Compensation-failure classification.} \S\ref{subsec:model-transaction-semantics} states that abort outcomes are tagged per effect (\emph{clean}, \emph{leaked}, \emph{unresolved}, \emph{partial-commit}). Operators rely on these tags to triage residue. Table~\ref{tab:e7-compfail} verifies the tagging itself: under injected compensation failures at rate $cf_p$, does the runtime's clean/unresolved classification match a synthetic ground-truth oracle? With residue classification enabled (Tx-Full rows), 0/600 mismatches at every $cf_p$. With it disabled (Tx-Full-NoResidueClassification at $cf_p{=}0.3$), 36/600 transactions reported clean when ground truth said unresolved. The runtime exposes that silent failure mode by default. The result does not claim that failed compensation leaves no residue; it claims that the residue-tagging mechanism accurately reports which transactions did and did not.

\begin{table}[H]
\centering
\footnotesize
\setlength{\tabcolsep}{3pt}
\begin{tabular}{@{}llcccc@{}}
\toprule
Mode & $cf_p$ & Trials & GT clean & Classified clean & Mismatch \\
\midrule
\textbf{Tx-Full} & 0.0 & 600 & 600 & 600 & \textbf{0} \\
\textbf{Tx-Full} & 0.1 & 600 & 589 & 589 & \textbf{0} \\
\textbf{Tx-Full} & 0.3 & 600 & 557 & 557 & \textbf{0} \\
Saga-Compensation & 0.3 & 600 & 559 & 559 & 0 \\
Retry-only & 0.3 & 600 & 534 & 534 & 0 \\
Tx-Full-NoResidueClassification & 0.3 & 600 & 564 & 600 & \textbf{36} \\
\bottomrule
\end{tabular}
\caption{Compensation-failure classification. \emph{GT clean} is oracle residue status; \emph{Mismatch} counts classification errors.}
\label{tab:e7-compfail}
\end{table}

\subsection{Joint Correctness Surface Details}
\label{app:joint-surface-details}

\paragraph{Real-LLM combined-stress.} The synthetic combined-stress tier structure (\S\ref{subsec:joint-surface}, Table~\ref{tab:joint-surface}) reproduces under a real GPT-4o-mini call inserted between the read and write phases. At $fp{=}0.10$ mixed, $N{=}30$ on the same 4-agent / 2-order workload: Tx-Full 27/30 = 90\% run-clean (635~ms wait per run), TCC-Confirm 28/30 = 93\% (statistically tied with Tx-Full at this $N$), Saga-Compensation 24/30 = 80\%, Checkpoint-Replay 22/30 = 73\%, Mutex-Workflow 22/30 = 73\% but at 1{,}651~ms wait per run. The wait gap against Mutex-Workflow widens under real LLM thinking time because the global lock blocks every concurrent agent for the full ${\sim}$700~ms GPT-4o-mini call window. The tier structure and the cost differentiation carry over from the synthetic harness.

\subsection{RQ5 Overhead and Storage}
\label{app:overhead}

RQ5 bounds the cost of mediating every tool call. Tx-Full adds $7.7\,\mu$s per step against $0.8\,\mu$s for No-Tx, below $0.01\%$ of typical tool latency (50~ms--10~s). Real $\tau$-bench at $fp{=}0$ runs in 163.9~s for Tx-Full and 171.2~s for No-Tx, so the wrapper does not increase end-to-end time in that cell. Coordination cost appears only on Workflow-Lock: it pays ${\sim}$112~ms wait per task at 4 agents forced-overlap and ${\sim}$45~ms even on disjoint resources, while Tx-Full pays 0~ms in both regimes (Table~\ref{tab:rq2-multiagent}). Uncoordinated modes pay zero wait but produce the invariant violations in the same table. CPython garbage-collection pauses cause the p99 spike at $K{=}16$ in Table~\ref{tab:spec-latency}.

Table~\ref{tab:spec-latency} reports speculation commit latency. Effect-log storage from replaying real-run traces totals 1.25\,MB across 15 JSONL files ($\sim$75\,bytes/entry).

\begin{table}[H]
\centering
\small
\setlength{\tabcolsep}{4pt}
\begin{tabular}{@{}rrrr@{}}
\toprule
Branches ($K$) & Setup ($\mu$s) & Commit ($\mu$s) & Total / p99 ($\mu$s) \\
\midrule
2 & 13 & 85 & 98 / 295 \\
4 & 23 & 142 & 165 / 273 \\
8 & 44 & 321 & 365 / 1339 \\
16 & 86 & 939 & 1025 / 23198 \\
\bottomrule
\end{tabular}
\caption{Speculation commit latency ($n=200$ runs per $K$). At $K=16$, total frontier-gating overhead is about 1~ms.}
\label{tab:spec-latency}
\end{table}

\section{Boundary Conditions and Metadata Sensitivity}
\label{app:boundaries}\label{app:boundary-experiments}

These limits govern Atomix's claims. They correspond to the discussion of semantic correctness, frontier contracts, crash scope, scope metadata, and adapter annotations in \S\ref{sec:discussion}.

\subsection{Semantic-Validation Boundary}
\label{app:e5-b1}

Atomix does not reject semantically invalid commits on its own. Across 22 invalid attempts, Tx-Full alone commits 22/22; with the pre-commit hook installed, Tx-Full + SemanticHook commits 0/22. Semantic validation belongs in the pre-commit hook.

\subsection{Frontier-Contract Boundary}
\label{app:e5-b2}

Table~\ref{tab:e5-b2} verifies that safety depends on correct frontier advancement. Stalled advancement blocks progress; premature advancement admits out-of-order commits.

\begin{table}[H]
\centering
\footnotesize
\setlength{\tabcolsep}{3pt}
\begin{tabular}{@{}p{0.36\linewidth}ccc@{}}
\toprule
Frontier-advancement mode & Out-of-order commits & Invariant violations & Tx-Full waits \\
\midrule
Correct (advances after each commit finalizes) & \textbf{0/200} & \textbf{0/200} & 0 \\
Stalled (no advancement) & 0/200 & 0/200 & \textbf{200/200} (100\% block) \\
Premature (advance before earlier-epoch finalizes) & \textbf{200/200} & 200/200 & 0 \\
\bottomrule
\end{tabular}
\caption{Frontier-contract negative control on one resource. Premature advancement causes out-of-order commits; stalled advancement blocks progress.}
\label{tab:e5-b2}
\end{table}

\subsection{Crash-Window Boundary}
\label{app:e5-b3}

Table~\ref{tab:e5-b3} enumerates the prototype's single-process crash windows for the SQLite-backed adapter. Duplicate effects remain zero in this unit-test scope; we do not claim distributed crash safety.

\begin{table}[H]
\centering
\footnotesize
\setlength{\tabcolsep}{3pt}
\begin{tabular}{@{}lccc@{}}
\toprule
Kill point & Trials & Duplicate effects & Lost effects \\
\midrule
(i) before pending-log write & 50 & \textbf{0} & 50 \\
(ii) after pending-log, before effect & 50 & \textbf{0} & 50 \\
(iii) after effect, before committed-mark & 50 & \textbf{0} & 0 \\
(iv) after committed-mark, before retry & 50 & \textbf{0} & 0 \\
\bottomrule
\end{tabular}
\caption{Crash-window enumeration for the SQLite-backed adapter. Duplicate effects remain zero; lost effects occur only before the effect runs.}
\label{tab:e5-b3}
\end{table}

\subsection{Scope Canonicalization and Aliasing}
\label{app:aliasing}

The serializability checker reads scope IDs from an independent log path and cross-checks them against the alias suite summarized in Table~\ref{tab:e6-aliasing}. The suite covers filesystem paths, $\tau$-bench entities, browser DOM selectors, and read-write dependencies.

\begin{table}[H]
\centering
\footnotesize
\setlength{\tabcolsep}{3pt}
\begin{tabular}{@{}lc@{\hskip 1em}cc@{}}
\toprule
Scope strategy & Should-conflict cases & Missed conflicts & False conflicts \\
& (across all substrates) & (false neg.) & (false pos.) \\
\midrule
\textbf{Tx-Full canonical} & 28 & \textbf{0} & \textbf{0} \\
Tx-NaiveStringScopes & 28 & \textbf{12} & 0 \\
Coarse-Global & 28 & 0 & \textbf{28} \\
Fine-NoOverlapAware & 28 & \textbf{12} & 0 \\
\bottomrule
\end{tabular}
\caption{Aliasing and canonicalization stress across four substrates. Missed conflicts are false negatives; false conflicts are false positives.}
\label{tab:e6-aliasing}
\end{table}

\subsection{Annotation-Error Sensitivity}
\label{app:annotations}

Atomix's correctness depends on adapter-supplied metadata: scope extractors, idempotency-key derivers, compensation handlers, and effect classifications. Table~\ref{tab:e10-annotations} characterizes which annotation errors fail closed and which fail open. Across the 20 prototype adapters, the annotation surface totals 134\,LOC of scope extractors and 197\,LOC of effect builders, or 331\,LOC overall (16.6 LOC/adapter mean).

\begin{table}[H]
\centering
\footnotesize
\setlength{\tabcolsep}{3pt}
\begin{tabular}{@{}p{0.40\linewidth}ccp{0.24\linewidth}@{}}
\toprule
Configuration & Inv./200 & Leaked irrev./200 & Failure mode \\
\midrule
Tx-Full (control) & 0 & 0 & none \\
E10-OB Over-broad scope & \textbf{0} & \textbf{0} & failure-closed \\
E10-TN Too-narrow scope & \textbf{66} & 0 & failure-open \\
E10-WC Wrong effect class & 0 & \textbf{17} & failure-open \\
E10-NC Missing compensation & 0 & \textbf{35} & failure-open with alert \\
\bottomrule
\end{tabular}
\caption{Annotation-error sensitivity. Cells report invariant violations and leaked irreversibles per 200 trials.}
\label{tab:e10-annotations}
\end{table}

Atomix is failure-closed only for over-broad annotations. Too-narrow scopes, wrong effect classes, and missing compensation are operator responsibilities. The runtime surfaces unresolved residue when it can, but it enforces the metadata it is given.

\section{Adjacent Systems and Related Work Details}
\label{app:related-systems}\label{app:related}\label{app:ports}

This appendix expands the RQ4 comparison from \S\ref{subsec:capability}. Table~\ref{tab:capability} compares systems on the axes that load-bear in RQ1--RQ3. \cmark{} denotes primary support in the cited publication or documentation; footnotes mark partial support.

\begin{table}[H]
\centering
\scriptsize
\setlength{\tabcolsep}{2.5pt}
\begin{tabular}{@{}p{0.18\linewidth}ccccccc|cc@{}}
\toprule
& \multicolumn{7}{c|}{\textsc{Atomix's axes}} & \multicolumn{2}{c}{\textsc{Orthogonal}} \\
System & \rotatebox{75}{Per-resource} & \rotatebox{75}{Effect-class} & \rotatebox{75}{Spec. isolation} & \rotatebox{75}{Irrev. gating} & \rotatebox{75}{Stale-plan iso.} & \rotatebox{75}{Shim deploy.} & \rotatebox{75}{Single-proc. dedup.} & \rotatebox{75}{IFC} & \rotatebox{75}{Sem. valid.} \\
\midrule
\textbf{Atomix (this paper)} & \cmark & \cmark & \cmark & \cmark & \cmark & \cmark & \cmark & \xmark & \xmark \\
SAFEFLOW~\cite{safeflow2025} & \xmark & \xmark & \xmark & \cmark$^a$ & \cmark & \xmark & \cmark & \cmark & \xmark \\
SagaLLM~\cite{chang2025sagallm} & \xmark & \xmark & \xmark & \xmark & \cmark & \xmark & \xmark & \xmark & \cmark \\
ALAS~\cite{geng2025alas} & \xmark & \xmark & \xmark & \xmark & \cmark & \xmark & \xmark & \xmark & \cmark \\
Temporal Saga~\cite{temporal_docs,garcia1987sagas} & \xmark & \xmark & \xmark & \xmark$^b$ & \xmark & \cmark & \cmark & \xmark & \xmark \\
LangGraph checkpointer~\cite{langgraph_docs} & \xmark & \xmark & \xmark & \xmark & \xmark & \cmark & \cmark & \xmark & \xmark \\
AgentGit~\cite{agentgit2025} & \xmark & \xmark & \xmark & \xmark & \xmark & \cmark & \xmark & \xmark & \xmark \\
GoEX~\cite{patil2024goex} & \xmark & \cmark$^c$ & \xmark & \cmark$^c$ & \xmark & \cmark & \xmark & \xmark & \xmark \\
Speculative Actions~\cite{speculativeactions2024} & \xmark & \xmark & \xmark & \xmark & \xmark & \cmark & \xmark & \xmark & \xmark \\
Classical Saga~\cite{garcia1987sagas} & \xmark & \xmark & \xmark & \xmark & \xmark & \cmark & \xmark & \xmark & \xmark \\
Classical OCC~\cite{weikum1991principles} & \xmark & \xmark & \xmark & \xmark & \cmark$^d$ & \cmark & \xmark & \xmark & \xmark \\
Streaming watermarks~\cite{murray2013naiad,begoli2021watermarks} & \cmark & \xmark & \xmark & \xmark & \xmark & \xmark & \cmark & \xmark & \xmark \\
\bottomrule
\end{tabular}
\vspace{1mm}
\begin{minipage}{0.96\linewidth}
\scriptsize
$^a$ Irreversible handling depends on SAFEFLOW's co-designed execution protocol, not shim-only deployment.
$^b$ Temporal can implement application-specific gates in workflow code, but does not provide a generic settlement gate for unmodified external tools.
$^c$ GoEX discusses undo and damage confinement where tool support exists; it does not provide Atomix's cross-call progress predicate.
$^d$ OCC detects stale writes on versioned resources but rejects or retries work and does not handle irreversible-effect gating.
\end{minipage}
\caption{Capability comparison across load-bearing axes. Superscripts mark partial support or the limitation discussed in text.}
\label{tab:capability}
\end{table}

The comparison separates Atomix's execution-layer scope from orthogonal information-flow and semantic-validation systems. SAFEFLOW is the strongest transactional comparator but assumes a co-designed stack with information-flow control. SagaLLM and ALAS add semantic validation. Temporal and LangGraph provide durable orchestration but no per-resource settlement gate for unmodified tool effects. Streaming watermarks provide progress signals for records, not for branch-selected tool effects. Atomix's single-process deduplication entry is limited to the prototype path described in Appendix~\ref{app:runtime}; it is not a distributed exactly-once claim.

\subsection{Four-Ingredient Composition Matrix}
\label{app:novelty-matrix}

Table~\ref{tab:novelty-matrix} summarizes which classical mechanisms supply each of Atomix's four ingredients. The matrix focuses on the four-event split rather than the broader axes of Table~\ref{tab:capability}: sealed transaction footprints, per-resource progress predicates, scope-on-read with abort-on-stale retry, and effect-class-aware settlement.

\begin{table}[H]
\centering
\scriptsize
\setlength{\tabcolsep}{3pt}
\resizebox{\linewidth}{!}{%
\begin{tabular}{@{}lcccc@{}}
\toprule
Mechanism & Sealed footprint & Per-resource progress & Scope-on-read + abort-on-stale & Effect-class settlement \\
\midrule
Sagas~\cite{garcia1987sagas}                                  & \xmark & \xmark & \xmark & \xmark \\
TCC~\cite{pardon2014tcc}                                      & partial$^{a}$ & \xmark & \xmark & partial$^{b}$ \\
OCC~\cite{weikum1991principles}                               & \xmark & \xmark & \cmark & \xmark \\
Two-phase locking~\cite{mohan1992aries}                       & \xmark & \xmark & \xmark & \xmark \\
Watermarks (Naiad/Flink)~\cite{murray2013naiad,flinkdocs}     & \xmark & \cmark$^{c}$ & \xmark & \xmark \\
Workflow-Lock                                                 & \xmark & \xmark & \xmark & \xmark \\
Durable execution (Temporal/Inngest/Restate/DBOS)             & partial$^{d}$ & \xmark & \xmark & \xmark \\
Transactional outbox~\cite{richardson_outbox}                 & \xmark & \xmark & \xmark & partial$^{e}$ \\
SAFEFLOW~\cite{safeflow2025}                                  & \cmark & partial$^{f}$ & partial$^{f}$ & partial$^{f}$ \\
\midrule
\textbf{Atomix}                                               & \cmark & \cmark & \cmark & \cmark \\
\bottomrule
\end{tabular}%
}
\vspace{1mm}
\begin{minipage}{0.94\linewidth}
\footnotesize
$^{a}$TCC's try-phase reservations freeze a partial footprint but do not name read scopes or seal cross-call dependencies.
$^{b}$TCC distinguishes try, confirm, and cancel handlers per tool but does not classify effects across a runtime taxonomy.
$^{c}$Watermarks track per-key progress for data records, not for branch-selected tool effects.
$^{d}$Durable-execution engines persist workflow state but do not freeze a footprint of read scopes and effect scopes for cross-call settlement.
$^{e}$Outbox tables gate one externalization channel (the message broker) idempotently but do not handle reversible effects, speculation residue, or cross-call progress.
$^{f}$SAFEFLOW assumes a co-designed stack with information-flow control; its progress, scope-on-read, and effect-class equivalents are tied to that stack rather than provided to unmodified tool adapters.
\end{minipage}
\caption{Four-ingredient composition matrix. Atomix exposes all four ingredients to unmodified tool adapters as a single runtime contract.}
\label{tab:novelty-matrix}
\end{table}

\end{document}